\documentclass[final,3p,times,twocolumn]{elsarticle}
\journal{Pattern Recognition}

\usepackage{graphicx}
\usepackage{algorithm}
\usepackage{algpseudocode}
\usepackage{epstopdf}
\usepackage[utf8]{inputenc}

\usepackage{amssymb}
\usepackage{amsmath,bm}
\usepackage{url}
\usepackage{subcaption}
\usepackage{lineno}
\usepackage{xcolor}
\usepackage{ulem}
\usepackage{setspace}
\usepackage{eqnarray}
\usepackage{balance}

\begin{document}

\begin{frontmatter}

%% Title, authors and addresses

%\title{Hyperparameter Tuning in SVM using NOMAD and a Dynamic Stopping Criteria}
%Ortho-MADS with Nelder Mead and Variable Neighborhood Search  

%\title{SVM Hyperparameter Tuning with an Orthogonal Direction Deterministic Mesh Adaptive Direct Search Algorithm and a Dynamic Stopping Criterion}

%\title{A Novel Orthogonal Direction Deterministic Mesh Adaptive Direct Search Algorithm for SVM Hyperparameter Tuning}

\title{A Novel Orthogonal Direction Mesh Adaptive Direct Search Approach for SVM Hyperparameter Tuning}

%% use the tnoteref command within \title for footnotes;
%% use the tnotetext command for the associated footnote;
%% use the fnref command within \author or \address for footnotes;
%% use the fntext command for the associated footnote;
%% use the corref command within \author for corresponding author footnotes;
%% use the cortext command for the associated footnote;
%% use the ead command for the email address,
%% and the form \ead[url] for the home page:
%%
%% \title{Title\tnoteref{label1}}
%% \tnotetext[label1]{}
%% \author{Name\corref{cor1}\fnref{label2}}
%% \ead{email address}
%% \ead[url]{home page}
%% \fntext[label2]{}
%% \cortext[cor1]{}
%% \address{Address\fnref{label3}}
%% \fntext[label3]{}

%% use optional labels to link authors explicitly to addresses:
 \author[label1,label2]{Alexandre R. Mello\corref{}}\ead{alexandre.reeberg@posgrad.ufsc.br} \ead[url]{https://orcid.org/0000-0003-3130-5328}
 \author[label1]{Jonathan de Matos}
 \ead{jonathan.de-matos.1@ens.etsmtl.ca}
   \author[label2]{Marcelo R. Stemmer}
   \ead{marcelo.stemmer@ufsc.br}
\author[label3]{Alceu de S. Britto Jr.}
   \ead{alceu@ppgia.pucpr.br}
\author[label1]{Alessandro L. Koerich}
\ead{alessandro.koerich@etsmtl.ca}
 \address[label1]{École de Technologie Supérieure - Université du Québec 1100 Notre-Dame West, Montréal, QC, H3C 1K3, Canada.}
 \address[label2]{University of Santa Catarina Campus Reitor João David Ferreira Lima, s/n - Trindade, Florianópolis, SC, 88040-900, Brazil.}
 \address[label3]{Pontifícia Universidade Católica do Paraná, Curitiba, PR, 80215-901, Brazil.}
%%%%%%%%%%%%%%%%%%%%%%%%%%%%%%%%%%%%%%%%%%%%%%%%%%%%%%%%%%%%%%%%%%%%
\begin{abstract}
In this paper, we propose the use of a black-box optimization method called deterministic Mesh Adaptive Direct Search (MADS) algorithm with orthogonal directions (Ortho-MADS) for the selection of hyperparameters of Support Vector Machines with a Gaussian kernel. Different from most of the methods in the literature that exploit the properties of the data or attempt to minimize the accuracy of a validation dataset over the first quadrant of $(C, \gamma)$, the Ortho-MADS provides convergence proof. We present the MADS, followed by the Ortho-MADS, the dynamic stopping criterion defined by the MADS mesh size and two different search strategies (Nelder-Mead and Variable Neighborhood Search) that contribute to a competitive convergence rate as well as a mechanism to escape from undesired local minima. We have investigated the practical selection of hyperparameters for the Support Vector Machine with a Gaussian kernel, i.e., properly choose the hyperparameters $\gamma$ (bandwidth) and $C$ (trade-off) on several benchmark datasets. The experimental results have shown that the proposed approach for hyperparameter tuning consistently finds comparable or better solutions, when using a common configuration, than other methods. We have also evaluated the accuracy and the number of function evaluations of the Ortho-MADS with the Nelder-Mead search strategy and the Variable Neighborhood Search strategy using the mesh size as a stopping criterion, and we have achieved accuracy that no other method for hyperparameters optimization could reach.

%however, most of them lack convergence proof or do not handle well undesired local minima.
\end{abstract}
%%%%%%%%%%%%%%%%%%%%%%%%%%%%%%%%%%%%%%%%%%%%%%%%%%%%%%%%%%%%%%%%%%%%
\begin{keyword}
NOMAD \sep hyperparameter tuning \sep model selection \sep SVM \sep MADS \sep VNS
%%%%%%%%%%%%%%%%%%%%%%%%%%%%%%%%%%%%%%%%%%%%%%%%%%%%%%%%%%%%%%%%%%%%
%% keywords here, in the form: keyword \sep keyword
%% MSC codes here, in the form: \MSC code \sep code
%% or \MSC[2008] code \sep code (2000 is the default)
\end{keyword}
\end{frontmatter}
%%
%% Start line numbering here if you want
%%
%%\linenumbers
%%%%%%%%%%%%%%%%%%%%%%%%%%%%%%%%%%%%%%%%%%%%%%%%%%%%%%%%%%%%%%%%%%%%
\section{Introduction}
\label{section:intro}
%%%%%%%%%%%%%%%%%%%%%%%%%%%%%%%%%%%%%%%%%%%%%%%%%%%%%%%%%%%%%%%%%%%%
Support Vector Machine (SVM) \cite{Boser:1992,Cortes1995} is a popular algorithm used in machine learning for statistical pattern recognition tasks, originally designed for binary classification \cite{Chih-WeiHsu2002}. The so-called maximal margin classifier optimizes the bounds of the generalization error of linear machines by separating the data with a strategy that attempts to find the maximal margin hyperplane in an appropriately chosen kernel-introduced feature space \cite{Cristianini2000}. An alternative to increase the SVM flexibility and improve its performance is a nonlinear kernel function such as the radial basis function (RBF or Gaussian). The SVM with Gaussian kernel has two important hyperparameters that impact greatly in the performance of the learned model: the soft-margin $C$ and the kernel hyperparameter $\gamma$. Therefore, the application of the SVM with Gaussian kernel to a classification problem requires an appropriate selection of hyperparameters, called hyperparameter tuning or model selection. Given these two hyperparameters and a training dataset, an SVM solver can find a unique solution of the constrained quadratic optimization problem and return a classifier model. Unfortunately, there is no standard procedure for hyperparameter tuning, and a common approach is to split the dataset into training and validation set, and for each $C$ and $\gamma$ from a suitable set, select the pair that results in an SVM model that when trained on the training set has the lowest error rate over the corresponding validation set \cite{Wainer2017}. 

Black-box optimization (BBO) is the study of the design and analysis of algorithms that assume that the objective and/or constraint functions are given by a black-box \cite{audet2017}. Some of the most used methods to solve BBO problems are: (i) the naive methods such as the exhaustive search, grid search, and coordinate search; (ii) the heuristic methods, such as the genetic algorithm (and its variations) and the Nelder-Mead search \cite{nelder1965}; and (iii) the direct search algorithms, such as the Generalized Pattern Search (GPS) and the Mesh Adaptive Direct Search (MADS) \cite{audet2006}. The naive and the heuristic methods do not guarantee convergence, while the direct search methods combine a flexible framework with proof of convergence \cite{audet2017}.

%RELATED WORK HERE----------------------------------------------------------------
Considering the relationship between the data geometric structure in the feature space and the kernel function (that is not relevant to $C$), \citet{JianchengSun2010} determine $C$ and $\gamma$ separately by two stages. \citet{Chen2017} proposed another two-step procedure for efficient parameter selection by exploiting the geometry of the training data to select $\gamma$ directly via nearest neighborhood and choosing the $C$ value with an elbow method that finds the smallest $C$ leading to the highest possible validation accuracy. This approach reduces the number of candidate points to be checked, while maintaining comparable accuracy to other classical methods. \citet{Chung2003} and \citet{Gold2005} introduced the Bayesian Optimization (BO) approach for tuning the kernel hyperparameters, which constructs a probabilistic model to find the minimum function $f(x)$ on some bounded set $\mathcal{X}$, that exploits the model to make decisions about where in $\mathcal{X}$ to next evaluate the function while integrating out uncertainty \cite{snoek2012}. \citet{luigi2017} proposed the Bayesian Adaptive Direct Search (BADS) which combines a Bayesian optimization with the MADS framework via a local Gaussian surrogate process, implemented with a number of heuristics. The BADS' goal is fitting moderately expensive computational models, and it achieved state-of-the-art accuracy on computational neuroscience models (such as convolution neural networks). \citet{Chang2015} proposed a two-stage method for hyperparameter tuning. The first step consists of tuning the $\gamma$ using the generalization error of a k-NN classifier with the aim of maximizing the margins and extend the class separation. The second stage defines the value of $C$ by an analytic function obtained with the jackknife technique.

Although many efforts to properly tuning the hyperparameters of a SVM with Gaussian kernel have been made, most of the BBO-based methods lack convergence proof. The BBO problems may present a limited precision or may be corrupted by numerical noise, which leads to an invalid output. Considering a fixed starting point, a BBO algorithm may provide different outputs, and there are unreliable properties frequently encountered in real problems \cite{Audet2014}. Furthermore, most of BBO-based methods use the time or number of function evaluations as a stopping criterion, which creates an uncertainty on the achieved local minimum.

The NOMAD software (Nonlinear Optimization by MADS) \cite{Nomad,Le2011a} is a \textit{C++} implementation of the MADS algorithm which can efficiently explore a design space in search of better solutions for a large spectrum of BBO problems as described by \citet{Audet2018a}, and inspired by cases of success as \cite{luigi2017} and \cite{Audet2018a}. In this paper, we use the Ortho-MADS \cite{abramson2009} (that is a MADS improvement), with two different search strategies Nelder-Mead \cite{nelder1965} and the Variable Neighborhood Search (VNS) \cite{Audet2008}, to tune hyperparameters of a SVM with Gaussian kernel considering a dynamic stopping criterion. The proposed method relies on the MADS convergence properties and it combines different search strategies to reach the desired local minimum (that is set by the mesh size) and escape from undesired local minimum. The experimental results on benchmark datasets have shown that the proposed approach achieves state-of-the-art accuracy, besides presenting several interesting properties such as the guarantee of convergence and a dynamic stopping criterion. Furthermore, it also provides many other tools that aid to adjust the hyperparameters regarding the particularities of each application. 

%UPDATE this in the end
This paper is organized as follows: Section \ref{section:basicconcepts} introduces the basic definitions of SVM, notation and the formulation of a BBO problem. Section \ref{section:proposed} presents the proposed approach for hyperparameter tuning. In Section \ref{section:experimens}, beyond describing other hyperparameter optimization methods used as benchmarks, we present our experimental protocol and the observed results. The conclusions and perspectives for future research are stated in the last section.

%Cross validation justification (if needed)
% \citet{Duan2003} and \citet{Duarte2017} compared six different internal metrics methods (Xi-Alpha bound \cite{Joachims2003}, Generalized approach cross-validation \cite{Wahba2001}, Approximate span bound \cite{Vapnik2000a}, VC bound \cite{Vapnik1998}, Radius-margin bound \cite{Chung2003}, and Modified radius-margin bound \cite{Cristianini1998}) with a standard cross validation to select hyperparameters on five different datasets, and concluded that the cross validation results in better classifiers, i.e., have lower expected error on future data.
%Also on resampling -> \cite{Wainer2017}

%%%%%%%%%%%%%%%%%%%%%%%%%%%%%%%%%%%%%%%%%%%%%%%%%%%%%%%%%%%%%%%%%%%%
\section{Basic Concepts of SVM}
\label{section:basicconcepts}
%%%%%%%%%%%%%%%%%%%%%%%%%%%%%%%%%%%%%%%%%%%%%%%%%%%%%%%%%%%%%%%%%%%%
%For this paper we define the training data as $H=[\bm{h_1}\cdots\bm{h_n}\in \mathbb{R}^{d\times n}]$, where $d$ is the number of inputs and $n$ is the number of features, and has two possible labels $y_i=\pm 1$.
%Small value of gamma the model behaves like a linear SVM, and when gamma is too large the model is heavily influenced by each support vector. C value is specific to the current data in usage, where a small C gives a wider margin at cost of some misclassification, a large C results in the hard margin classifier and do not tolerate constraint violation. 
Let's consider a training dataset $H=[\bm{h_1}\cdots\bm{h_n}\in \mathbb{R}^{d\times n}]$, where each $h_i$ consists of $n$ features and one possible class label $y_i=\pm 1$. We can reduce the SVM problem to a convex optimization form, i.e., minimize a quadratic function under linear inequality constraints\footnote{All proves and deductions can be found in \cite{Cristianini2000}.}. Given a kernel function $\kappa (\bm{h},\bm{h'})$ that must satisfy the distance relationship between transformed and original space, i.e., the kernel function satisfies $\kappa (\bm{h},\bm{h'})= \Phi(\bm{h})^\top \Phi(\bm{h'})$, it can map the training instances into some feature space $\mathcal{F}:\mathbb{R}^d \rightarrow \mathcal{F}$. The Gaussian Radial Basis Function (RBF) kernel is a usual choice for the kernel function, and it is defined as:

\begin{equation}
\begin{aligned}
\kappa (\bm{h}, \bm{h^\prime}) & := \exp(\frac{-||\bm{h}-\bm{h'}||^2}{2 \sigma^2}) \\
& \;\;=\exp({\gamma ||\bm{h}-\bm{h'}||^2}), \qquad \forall \bm{h},\bm{h'} \in \mathbb{R}^d
\end{aligned}
\label{eq:rbf}
\end{equation}

\noindent where $||\cdot ||$ represents the $l_2$ norm and $(\gamma,\sigma)$ are fixed constants with $\gamma=\frac{1}{2\sigma^2}$. The maximum margin separating hyperplane in the feature space is defined as $\bm{\omega}^\top\Phi(\bm{h})+b=0$, and the quadratic programming problem that represents the nonlinear soft-margin SVM is described by Eq.~\ref{eq:SVM} in its primal form. The minimization of Eq.~\ref{eq:SVM} results in the maximum margin separating hyperplane. 

\begin{equation}\label{eq:SVM}
\begin{aligned}
\min_{\bm{\omega},b,\xi} \quad & \frac{1}{2}||\bm{\omega}||^2 + C \sum_{i=1}^{n} \eta_i \\
\text{subject to  } &y_i(\bm{\omega} \cdot \Phi(\bm{h_i}) + b) \geq 1 - \eta_i\\
& \eta_i \geq 0, \forall i
\end{aligned}
\end{equation}

\noindent where $\eta$ is the slack variable that indicates tolerance of misclassification, the constant $C>0$ is a trade-off hyperparameter. We calculate the classification score using Eq.~\ref{eq:classification}, regarding the distance from an observation $\bm{h}$ to the decision boundary.

\begin{equation}\label{eq:classification}
f(h)=\sum_{i=1}^d \bm{\alpha}_i y_i \kappa(\bm{h}_i,\bm{h})+b
\end{equation}

\noindent where $\bm{\alpha}_1,\cdots,\bm{\alpha}_d$ and $b$ are the estimated SVM parameters. For the cases where we have a multiclass problem, we can extend the SVM by reducing the classification problem with three or more classes to a set of binary classifiers using the one-vs-one or the one-vs-rest approach \cite{Chih-WeiHsu2002}. In this paper, we use the one-vs-one error-correcting output codes (ECOC) model \cite{Dietterich1994}. Although, the choice of the multiclass approach does not interfere in our proposed solution.

A practical difficulty in the nonlinear SVM is to proper tune the hyperparameters $\gamma$ and $C$ from Eqs.~\ref{eq:rbf} and \ref{eq:SVM} respectively. For the BBO-based methods (that are not based on internal metrics), the tuning process maximizes the accuracy on a training data usually using a cross-validation procedure, which is equivalent to minimizing a loss function ($\mathcal{L}$), as described by Eq.~\ref{eq:Accuracy}.

%\begin{equation}\label{eq:Accuracy}
%\max_{\gamma>0,\; C>0} \; \text{Accuracy}(\gamma,C)\quad \equiv \quad \min_{\gamma>0,\;  C>0} \; L (\gamma,C)
%\end{equation}

\begin{equation}\label{eq:Accuracy}
\max_{\gamma>0,\; C>0} \; \text{Accuracy}(\gamma,C)\quad \equiv \quad \min \; \mathcal{L}
\end{equation}

Almost all loss functions commonly used in the literature met convexity assumption, however, different loss functions lead to different theoretical behaviors, i.e., different convergence rates \cite{Rosasco2004}. It is possible to use any loss function that meets convexity with the proposed approach, but based on \citet{Rosasco2004} and considering the loss function options for classification tasks, we have chosen the Hinge loss (HL). The HL leads to a convergence rate practically equal to the convergence rate of a logistic loss and better than the convergence rate of a square loss. However, considering a hypothesis space sufficiently rich, the thresholding stage has little impact on the obtained bounds. We define the weighted average classification hinge loss to tune the hyperparameters of the SVM as Eq.~\ref{eq:Accuracy2}.

\begin{equation}\label{eq:Accuracy2}
\mathcal{L}=\sum_{i=1}^d w_i \max \{ 0.1-m_i \}
\end{equation}
 
\noindent where $w_i$ is the normalized weight for an observation $i$ ($\sum_{i=1}^{d} w_i=1$) and $m_i = y_i f(\bm{h_i})$ is the scalar classification score when the model predicts true for the instance $\bm{h_i}$. 

% $f(\bm{h_i})$ is the predictor data and $\bm{h}$ is the raw classification score for an observation $i$.

% Moved from the introduction to here:
 We can formulate the SVM with Gaussian kernel hyperparameter tuning problem as a BBO function $f(x): \mathbb{R}^n\rightarrow\mathbb{R}$ in the sense that we obtain a function value from a given $x\in\mathbb{R}$ (in this case $x$ corresponds to the hyperparameters $C$ and $\gamma$) for which the analytic form of $f$ is unknown. The objective function $f$ and the different functions defining the set $\Omega$ are provided as a bounded optimization problem of the form:

\begin{equation}\label{eq:fx}
\min_{z\;\in\;\Omega\subseteq\mathbb{R}^n} f(x)
\end{equation}

\noindent where $f:\mathbb{R}^n\rightarrow\mathbb{R}\cup\{\infty\}, \; \Omega=\{x\in X : c_j(x)\leq 0, j=1,2,\cdots,m\} $, and $X\in\mathbb{R}^n$ represents closed constraints which are bounded by $L\leq x\leq U$, with the lower ($L$) and the upper ($U$) bounds in $(\mathbb{R}\cup\{\pm\infty\})^n$, and the functions $c_j$ represent the other $m$ open constraints \cite{Audet2008,Zegal2012}.

\section{The Ortho-MADS with NM and VNS}
\label{section:proposed}
%%%%%%%%%%%%%%%%%%%%%%%%%%%%%%%%%%%%%%%%%%%%%%%%%%%%%%%%%%%%%%%%%%%%
In this section, we briefly introduce the methods that we propose for tuning the hyperparameters of a SVM with Gaussian kernel, which are the MADS and the \textit{Ortho} variation, as well as the two search methods used with the Ortho-MADS: the Nelder-Mead (NM) search and the Variable Neighborhood Search (VNS). 
%%%%%%%%%%%%%%%%%%%%%%%%%%%%%%%%%%%%%%%%%%
\subsection{Mesh Adaptive Direct Search (MADS)}
%%%%%%%%%%%%%%%%%%%%%%%%%%%%%%%%%%%%%%%%%%
The Mesh Adaptive Direct Search (MADS) \cite{audet2006} is an iterative algorithm to solve problems of the form Eq.~\ref{eq:fx}, in which the goal at each iteration is to replace the current best feasible points (called incumbent solution, or the incumbent) by a better one. The algorithm starts from a finite collection of initial points $V_0\in\mathbb{R}^n$, and the incumbent $x_k$ at iteration $k$ is defined as $f(x):x\in V^k\cap\Omega$, where $V^k$ is the set of trial points where the black-box was previously evaluated. To achieve a better solution at each iteration it uses a mesh $M^k$ to generate trial points on a discretized space, defined by Eq.~\ref{eq:mesh}.
%solving nonlinear unconstrained optimization problems. The idea is to create a mesh around a point that provides best results for the objective function. Eq.~\ref{eq:mesh} defines the mesh.

%The MADS algorithm will generalize the GPS algorithm by allowing
%an infinite set of polling directions, while improving the convergence results
\begin{equation}\label{eq:mesh}
M^k := \{x^k + \delta^k D \; y : y \in \mathbb{N}^p\}
\end{equation}

\noindent where $\delta^k$ defines the mesh size, $D$ is the set of directions, and $p$ is the number of directions. At iteration $k$, the algorithm attempts to improve the incumbent solution by executing the \textit{search} or \textit{poll} stage, allowing an attempt of improving the current solution or escaping from an undesired local minimum. The \textit{search} stage attempts to improve the incumbent by evaluating points that are generated from a finite trial points of the subset $M^k$, based on the $D$ set of directions, and that are not necessarily close to the current incumbent. When the \textit{search} stage does not succeed, the \textit{poll} stage is executed to evaluate points inside a frame defined by Eq.~\ref{eq:eqframe}.  

\begin{equation}
F^k := \{x \in M^k :  \parallel x - x^k \parallel_\infty < \Delta^kb\}
\label{eq:eqframe}
\end{equation}

\noindent where $x$ is the point being evaluated, $\Delta^k$ is the frame size that must satisfy $0<\delta^k \leq\Delta^k$, and $b$ is defined by Eq.~\ref{eq:eqbframe}.

\begin{equation}
b = \max\{\parallel d^\prime \parallel_\infty : d^\prime \in D\}
\label{eq:eqbframe}
\end{equation}

\noindent where $d^\prime$ is a direction.

The frame is always larger than the mesh and provides a broader sampling. If the \textit{poll} stage obtains a new incumbent, the frame moves to this point and increases its size, otherwise, it decreases its size and maintains the current incumbent. We define the initial poll size as $\Delta^0_j=\frac{U_j-L_j}{10}$ according to \cite{Audet2016}, and the initial mesh size as $\delta^0=\Delta^0$. The \textit{poll} stage selects points based on a subset $D^k$ of directions from $D$, $\delta$ and $\Delta$, and at each iteration, the mesh size $\delta$ is updated as $\delta=min\{\Delta^k,{\Delta^k}^2\}$, changing the frame size by $\tau^{-1}\Delta$ when increased and $\tau^{1}\Delta$ when decreased, where $\tau\in [0,1]$. This process allows a continuous refinement of the mesh around the incumbent, and consequently reducing the objective function value. 

The Ortho-MADS \cite{abramson2009} is an improvement in the \textit{poll} stage of the MADS algorithm, where the choice of the polling directions is deterministic and orthogonal to each other. This leads, at each iteration, to convex cones of missed directions that are minimal in a reasonable measure \cite{abramson2009}. The Ortho-MADS changes the evaluation point selection method during the \textit{poll} stage by creating $D^k$ set of orthogonal directions, which define a dense sphere that increases with the number of iterations $k$\footnote{for further information refers to \cite{abramson2009}.}.

\begin{figure*}[htpb!]
\centering
\subcaptionbox{\label{fig:mesh1}}{\includegraphics[width=0.40\textwidth]{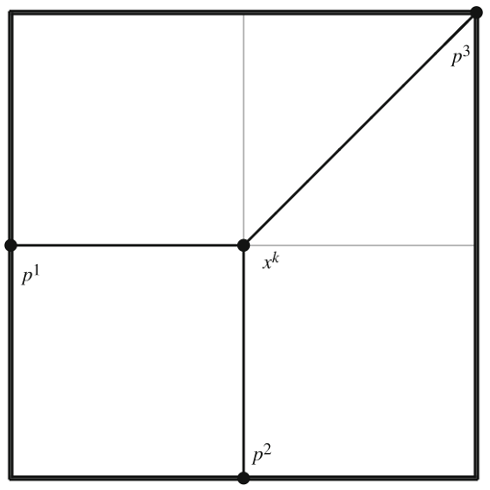}}
\hspace{30pt}
\subcaptionbox{\label{fig:mesh2}}{\includegraphics[width=0.40\textwidth]{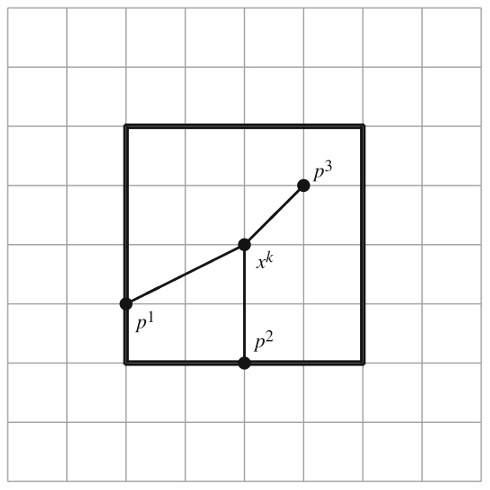}}
\label{fig:mesh}
\caption{Two mesh and frame representations with different sizes: (a) mesh size $\delta^k=1$ and frame size $\Delta^k=1$; (b) mesh size $\delta^k=\frac{1}{4}$ and frame size $\Delta^k=\frac{1}{2}$ (Images from \cite{audet2017})}
\end{figure*}

Fig.~\ref{fig:mesh1} depicts a mesh and frame of size one, where the \textit{poll} stage points $p^1$, $p^2$, and $p^3$ are limited by the frame and the mesh size, which in this case, the \textit{search} and \textit{poll} stage would find both the same points, performing similarly. In Fig.~\ref{fig:mesh2}, the mesh size is half of the frame size, and the \textit{poll} stage can use 24 points defined by the mesh intersections (excluding $x^k$), allowing a broader exploration of the search space than the searching step.

%HERE
Among the traditional stopping criteria, such as time and number of function evaluations, the MADS can terminate the optimization process when it achieves a desired minimum mesh size value, that corresponds to achieve a desired local minimum. During the MADS optimization process, it continuously updates the mesh ($\delta$) and the frame ($\Delta$) sizes. Both $\delta$ and $\Delta$ become smaller as the algorithm moves towards the minimum of $f(x)$. This mesh size verification is the last step in the MADS iterations. For the SVM with Gaussian kernel, we have two hyperparameters (thus two variables) $C$ and $\gamma$ that may have different scales, e.g. $C\in [0, 100]$ and $\gamma\in [1e^{-5}, 1]$. The NOMAD can assume different minimum mesh sizes for each dimension, pre-defined at the beginning of the BBO optimization process.

%This stop criterion is an advantage comparing to other hyperparameter search mechanisms which do not provide a stop criterion by convergence, only for iterations number.

%%%%%%%%%%%%%%%%%%%%%%%%%%%%%%%%%%%%%%%%%%
\subsection{Search Methods}
%%%%%%%%%%%%%%%%%%%%%%%%%%%%%%%%%%%%%%%%%%
The \textit{search} stage is not necessary for the convergence analysis and can be done using different strategies such as the Variable Neighborhood Search (VNS) \cite{Audet2008} or the Nelder-Mead (NM) search \cite{nelder1965}. The MADS algorithm does not dictate the selection of points in the \textit{search} stage. 

The NM \cite{nelder1965} is a method for function minimization proposed by \citet{audet2018} to be used in the \textit{search} stage of the MADS algorithm. The NM continuously replaces the worst point $x^n$ from a set of $n$ points defined as $\mathbb{X} = \{x^0, x^1, \cdots, x^n\}$, where $\mathbb{X}$ is a set of $n$ vertices from a simplex problem for the minimization function $f(x)$. A point $x_{new}$ is considered better, or is said to dominate another point $x$, if $f(x_{new})<f(x)$. The domination defines the function Best$(x, x_{new})$ as in Eq.~\ref{eq:best}.

\begin{equation}
\text{Best}(x_{new},x) = \begin{cases}x_{new}, & \text{if $f(x_{new}) < f(x)$} \\
x, & \text{otherwise}
\end{cases}
\label{eq:best}
\end{equation}

At each iteration, the NM evaluates $f(x)$ at the points given by the simplex and replaces $x^n$ according to the following criteria:

	\begin{eqnarray}
	x^n=
    \begin{aligned}
    \begin{cases}
    shrink(\mathbb{X}) & \text{if $x^{ic}\in$ inside contraction zone}\\
    x^{ic} & \text{if $x^{ic}\notin$ inside contraction zone}\\
    \text{Best}(x^r, x^e) & \text{if $x^r\in$ expansion zone}\\
    x^r & \text{If $x^r\in$ reflexion zone}\\
    \text{Best}(x^r, x^{oc}) & \text{If $x^r\in$ outside contraction zone}\\
    \end{cases}
	\end{aligned}
	\end{eqnarray}

%{\color{red}
%VERIFICAR E ELIMINAR 
%\begin{itemize}
%  \setlength{\parskip}{0pt}
%  \setlength{\itemsep}{0pt plus 1pt}
%\item If $x^r$ belongs to inside contraction zone
%\begin{itemize}
%  \setlength{\parskip}{0pt}
%  \setlength{\itemsep}{0pt plus 1pt}
%\item If $x^{ic}$ belongs to the inside contraction zone, then $shrink(\mathbb{X})$
%\item else $x^n = x^{ic}$
%\end{itemize}
%\item If $x^r$ belongs to the expansion zone, $x^n = \text{Best}(x^r, x^e)$
%\item If $x^r$ belongs to the reflexion zone, $x^n = x^r$
%\item If $x^r$ belongs to the outside contraction zone, $x^n = \text{Best}(x^r, x^{oc})$
%\end{itemize}
%}

\noindent where $x^r$, $x^e$, $x^{oc}$, and $x^{ic}$ are defined by Eqs.~\ref{eq:nm2} to \ref{eq:nm5} as follows:

\begin{equation}
\label{eq:nm1}
x^c = \frac{1}{n}\sum_{i=0}^{n-1}x^i
\end{equation}

\begin{equation}
\label{eq:nm2}
x^r = x^c + (x^c - x^n)
\end{equation}

\begin{equation}
\label{eq:nm3}
x^e = x^c Q^e(x^c - x^n)
\end{equation}

\begin{equation}
\label{eq:nm4}
x^{oc} = x^c Q^{oc}(x^c - x^n)
\end{equation}

\begin{equation}
\label{eq:nm5}
x^{ic} = x^c Q^{ic}(x^c - x^n)
\end{equation}

\begin{equation}
%\label{eq:nm6}
\begin{aligned}
\text{shrink} (\mathbb{X}) = & x^0, x^0+\zeta(x^1-x^0), x^0+\zeta(x^2-x^0),\\ 
& \cdots, x^0+\zeta(x^n-x^0)
\end{aligned}
\end{equation}

\noindent where $Q^e$, $Q^{oc}$, and $Q^{ic}$ are expansion, outside contraction and inside contraction respectively, being defined as $Q^e=2$, $Q^{oc}=-\frac{1}{2}$, and $Q^{ic}=\frac{1}{2}$.  $\zeta$ is the shrinking parameter usually defined as $\zeta^{ic}=\frac{1}{2}$.

The zone definition of a new point $x$ is given as follows:

\begin{equation*}
\small
x\in
    \begin{aligned}
    \begin{cases}
    \text{inside contraction zone} & \text{if $x^n$ dominates $x$}\\
    \text{expansion zone} & \text{if $x$ dominates $x^0$}\\
    \text{reflection zone} & \text{if $x$ dominates at least two $\mathbb{X}$ } points\\
    \text{outside contraction zone} & \text{otherwise}
    \end{cases}
    \end{aligned}
\end{equation*}

\noindent and in the last case, $x$ dominates none or one point of $\mathbb{X}$.

%\begin{itemize}
%  \setlength{\parskip}{0pt}
%  \setlength{\itemsep}{0pt plus 1pt}
%\item Belongs to the inside contraction zone if $x^n$ dominates $x$
%\item Else, it belongs to the expansion zone if $x$ dominates $x^0$
%\item Else, it belongs to the reflection zone if $x$ dominates at least 2 points of $\mathbb{X}$
%\item Else, it belongs to the outside contraction zone and $x$ dominates 0 or 1 point of $\mathbb{X}$
%\end{itemize}

The NM method improves the search stage by selecting new points in a more controlled way than e.g., a random selection. The MADS algorithm can also use other methods in the search stage, such as the Bayesian Optimization method, which is the basis of the BADS algorithm. The convergence rate of the MADS algorithm depends on the quality of the search method. 

%\subsection{Variable Neighborhood Search (VNS)}
In order to include a far-reaching search step to escape from an undesired local minimum, \citet{Audet2008} also incorporates the Variable Neighborhood Search (VNS) \cite{Mladenovic1997,Hansen2001} as a \textit{search} stage in the MADS algorithm. The VNS algorithm complements the MADS \textit{poll} stage, i.e., when current iteration results in no success, which means that it could not find points with a smaller $f(x)$, the next \textit{poll} stage generates trial points closer to the poll center, while the VNS explores a more distant region with a larger perturbation amplitude. The VNS uses a random perturbation method to attempt to escape from a local optimum solution so that a new descent method from the perturbed point leads to an improved local optimum. The VNS requires a neighborhood structure that defines all possible trial points reachable from the current solution, and a descent method that acts in the structure. The VNS amplitude of iteration $k$ is parameterized by a non-negative scalar $\xi_k\in \mathbb{N}$ that gives the order of the perturbation.

The MADS mesh provides the required neighborhood structure to the VNS, and by adding a VNS exploration in the search step, it introduces two new parameters: one is related to the VNS shaking method $\Delta_v>0$, and the other defines a stopping criterion for the descent $\rho>0$ \cite{Audet2008}. The shaking of iteration $k$ generates a point $x^\prime$ belonging to the current mesh $M(k, \Delta_k)$, and the amplitude of the perturbation is relative to a coarser mesh, which is independent of $\Delta_k$. The VNS mesh size parameter defined as $\Delta_v>0$ and the VNS mesh $\text{M}(k,\Delta_v)$ are constant and independent on the iteration number $k$ not to be influenced by a specific MADS behavior (the perturbation amplitude $\xi_k$ is updated outside the VNS search step). The shaking function is defined as:

\begin{equation}
\begin{aligned}
\text{shaking}: (M(k, \Delta_k), \mathbb{N}) \rightarrow M(k, \Delta_v) \subseteq M(k, \Delta_k) \\
(x, \xi_k) \longmapsto x^\prime = \text{shaking}(x, \xi_k)
\end{aligned}
\end{equation}

The VNS descent function generates a finite number of mesh points and it is defined as:

\begin{equation}
\begin{aligned}
\text{descent}:M(k, \Delta_v) \rightarrow M(k, \Delta_k) \\
x^\prime \longmapsto x^{\prime\prime} = \text{descent}(x^\prime)
\end{aligned}
\end{equation} 

\noindent where $x^\prime$ is the point resultant from the previous shaking and $x^{\prime\prime}$ is a point based on $x^\prime$ but with improved $f(x)$ value. The improvement is important because $x^\prime$ has low probability to generate good optimization results due to its random choice by the shaking.

The descent step must lead towards a local optimum, and in the MADS context the local optimality is defined with respect to the mesh, so the descent step acts with respect to the current step size $\Delta_k$ and the directions used. To reduce the number of function evaluations and to avoid exploring a previously visited region, the descent step stopping criterion is defined as $||x-x_{new}||_\infty\leq\rho$, where $x_{new}$ is a trial point close to another point $x$ considered previously\footnote{Further information of the VNS and the MADS integration can be found at \cite{Audet2008}}. The VNS generally brings about a higher number of black-box evaluations, but these additional evaluations lead to better results \cite{Audet2008}.

%We propose to use the Ortho-MADS with NM and VNS as search stage to tune the hyperparameters of a Gaussian SVM using the mesh size $(\delta^k)$ as a dynamic stopping criteria.

We choose to use the mesh size parameter $\delta_k$ as stopping criterion because it corresponds to a situation where new refinements could not find a better solution, meaning a local minimum, and according to \citet{audet2017}, the mesh size parameter goes to zero faster than the poll size parameter $\Delta_k$. The NM improves the quality of the solutions in the \textit{search} stage \cite{audet2018}, while the VNS allows the far-reaching exploration from current incumbent \cite{Audet2008}. The pseudo-code in Alg.~\ref{alg:NOMAD} describes the high-level procedure of the MADS technique considering the Ortho-MADS algorithm with the Nelder-Mead and the VNS search strategies and the minimum mesh size as stopping criterion implemented by NOMAD.

To run the NOMAD software with the Ortho-MADS, NM, and VNS, we need to set the lower bound $L=[l_1, l_2]$, the upper bound $U=[u_1,u_2]$, and the initial poll and mesh size are defined as $\Delta^0_j=\delta^0_j=\frac{u_j-l_j}{10}$, where the mesh size parameter is associated with the variable $j\in\{1, 2, \dots, n\}$. The algorithm (Alg.~\ref{alg:NOMAD}) starts evaluating $f(x)$ within the initial point. The first iteration executes a search step, and in case of a failure, that is, not finding a $f(x)$ value smaller than $f(x_k)$, it executes a poll step with Ortho-MADS direction. In case of a successful iteration of the NM-Search and the poll step, the next iteration runs the VNS-search to try escaping from an eventual and undesired local minimum.

\begin{algorithm}[htpb!]
\footnotesize
		\caption{NOMAD high-level procedure}\label{alg:NOMAD}
		\textbf{Input:} Initial point $x_0=\{C_0,\gamma_0 \}$, VNS amplitude parameter $\xi$, Minimum mesh size $\{\delta_{\min C},\delta_{\min \gamma} \}$ \\
		\textbf{Output:} Best point $x_{\text{best}}=\{C_{\text{best}},\gamma_{\text{best}}\}$  \\
		\textbf{Initialization:} $k \gets 0$
		\begin{algorithmic}[1]
			\While{$\delta^{k}_{C}>\delta_{\min C} \text{\textbf{ and }} \delta^{k}_{\gamma}>\delta_{\min \gamma}$}
			\State{$x^{\prime} \leftarrow shaking(x_k, \delta_k)$}
            \State{$x^{\prime\prime} \leftarrow descent(x^{\prime})$}
            \renewcommand{\algorithmicrequire}{\textbf{\ \ \ \ \ \ \ \ Search stage}}
            \Require
            \State{$S_k \leftarrow$ finite number of points of $M(k,\Delta_k)$}
            \Comment{$M(k,\Delta_k)$ is the current mesh size}
            \State{Evaluates $f(t)$ on $S_k \cup x^{\prime\prime}$}
            \Comment{$f(t)$ is the sub-step evaluation inside the search stage}
			\If{$f(t) < f(x_k)$ for some $t$ in a finite subset of $S_k \subset M_k$ using NM-SEARCH}
			    \State{$x_{k+1} \gets t$}
			    \State{\textbf{goto} 20}
			\EndIf
	        \renewcommand{\algorithmicrequire}{\textbf{\ \ \ \ \ \ \ \ Poll stage}}
            \Require
			\State{Compute $p$ MADS directions $D_k \in \mathbb{R}^{n x p}$}
			\Comment{$p$ is the number of Poll stage points and $D_k$ is the directions set at $k$}
			\State{Construct the frame $P_k \subseteq M(k,\Delta_k)$}
			\State{Evaluates $f(p)$ on $p$ points of $P_k$}
			\If{$f(p) < f(x_k)$}
			    \State{$x_{k+1} \gets t$}
			    \State{$\Delta_{k+1} \gets \tau^{-1}\Delta_k$}
			\Else
			    \State{$x_{k+1} \gets x_k$}
			    \State{$\Delta_{k+1} \gets \tau^{1}\Delta_k$}
			\EndIf
			\State{Update VNS amplitude ($\xi_{k+1} \gets \xi_0 \text{ or } \xi_{k+1} \gets \xi_k+\delta$)}
			\State{Updates of solution and mesh}
			\State{$k \gets k+1$}
			\EndWhile \\
			\Return $x_k$
			%{\color{red}\Return ?????}
		\end{algorithmic}
% 		\begin{algorithmic}[1] 
% 		\textbf{Initialization}
% 		    $k \leftarrow 0$
% 		\textbf{Poll and search step}
% 		    \textbf{SEARCH step:}
% 		        $x\prime \leftarrow shaking(x_k, \delta_k)$
% 		        $x \prime\prime \leftarrow descent(x\prime)$
% 		        $S_k \leftarrow finite number of points of M(k,\Delta_k)$
% 		        evaluates f(x) on $S_k$
% 		\end{algorithmic}
\end{algorithm}

%end of color red
%%
%% Experiment
%%
%%%%%%%%%%%%%%%%%%%%%%%%%%%%%%%%%%%%%%%%%%
\section{Experimental Protocol and Results}
\label{section:experimens}
%%%%%%%%%%%%%%%%%%%%%%%%%%%%%%%%%%%%%%%%%%
We use the NOMAD black-box optimization software \cite{Nomad,Le2011a} (version 3.9.1), which provides several interfaces to run the Ortho-MADS and its variations, including MATLAB. The other approaches used for comparison are also implemented in MATLAB, and we developed an experimental protocol to compare the black-box optimization methods in a machine learning classification context based on \citet{audet2017} and \citet{More2009}. The experimental protocol consists of three steps:
	\begin{enumerate}
		\item Select datasets;
        \item Algorithm comparison using a common configuration;
		\item Evaluation of the proposed strategy.
	\end{enumerate}
    
%%%%%%%%%%%%%%%%%%%%%%%%%%%%%%%%%%%%%%%%%%
\subsection{Datasets}
%%%%%%%%%%%%%%%%%%%%%%%%%%%%%%%%%%%%%%%%%%
We have selected thirteen benchmark datasets with different numbers of instances and dimensions to evaluate the proposed approach, and to compare it with other strategies available in the literature. These datasets are publicly available at the LIBSVM website\footnote{\url{https://www.csie.ntu.edu.tw/∼cjlin/libsvmtools/datasets/} }. Tab.~\ref{tab:datasets} summarizes the main characteristics of each dataset.

\begin{table}[htpb!]
\footnotesize
\centering
\begin{tabular}{l r r r r r r}
\hline
\textbf{Dataset} & \textbf{\#Class} & \textbf{\#Features} & \textbf{\#Train} & \textbf{\#Test} & \textbf{\#Valid} & \\
\hline
Astroparticle \cite{wei2003} & 2 & 4 & 3,089 & 4,000 & NA\\
Car \cite{wei2003} & 2 & 21 & 1,243 & 41 & NA\\
DNA \cite{Chih-WeiHsu2002} & 3 & 180 & 1,400 & 1,186 & 600 \\
Letter \cite{dua2017} & 26 & 16 & 10,500 & 5,000 & 4,500 \\
Madelon \cite{guyon2005} & 2 & 500 & 2,000 & 600 & NA\\
Pendigits \cite{dua2017} & 10 & 16 & 7,494 & 3,498 & NA\\
Protein \cite{wang2003} & 3 & 357 & 14,895 & 6,621 & 2871 \\
Satimage \cite{Chih-WeiHsu2002} & 6 & 36 & 3,104 & 2,000 & 1,331 \\
Shuttle \cite{Chih-WeiHsu2002} & 7 & 9 & 30,450 & 14,500 & 13,050 \\
Splice \cite{dua2017} & 2 & 60 & 1,000 & 2,175 & NA\\
SVMguide4 \cite{wei2003} & 4 & 10 & 300 & 312 & NA\\
USPS \cite{hull1994} & 10 & 256 & 7,291 & 2,007 & NA \\
Vowels \cite{dua2017} & 11 & 9 & 598 & 462 & NA\\
\hline
{\scriptsize{NA: Not available.}}
\end{tabular}
\caption{Benchmark datasets used in the experiments.}
\label{tab:datasets}
\end{table}

%%%%%%%%%%%%%%%%%%%%%%%%%%%%%%%%%%%%%%%%%%
\subsection{Other BBO methods}
\label{section:alg_comparison}
%%%%%%%%%%%%%%%%%%%%%%%%%%%%%%%%%%%%%%%%%%

We have selected five widely used BBO-based methods of hyperparameter tuning for comparison purposes: Bayesian optimization (BO), Bayesian Adaptive Direct Search (BADS), Simulated Annealing (SA), Grid Search (GS), and Random Search (RS). These methods are briefly described as follows:  

\begin{itemize}
\item The Bayesian optimization (BO) \cite{snoek2012} uses a probabilistic function $f(x)$ as a model for the problem. It benefits from previous information in contrast with other methods that use gradients or Hessians. Bayesian optimization uses prior, which is the probabilistic model of the objective function, and the acquisition function, which defines the next points to evaluate. In this comparison, we used the Gaussian process as prior and the {\it expected improvement} as acquisition function. We use the \textit{bayesopt} function from MATLAB.
\item The Bayesian Adaptive Direct Search (BADS) \cite{luigi2017} uses a Mesh Adaptive Direct Search (MADS) \cite{audet2006} with a Bayesian optimization in the search step. The search step performs the discovering of the points with the intention of inserting domain-specific information and improving the quality of the points. When the search stage does not find suitable points, the poll stage of MADS broadly evaluates new points. The poll step is a computational expensive process and explores the objective function's shape for new points. We use the MATLAB implementation provided by \citet{luigi2017}\footnote{Available at \url{https://github.com/lacerbi/bads}}.
\item The Simulated Annealing (SA) mimics the process of annealing in metal. The temperature value dictates the probability function of the distance to a new random point based on the current point, while the distance to new points reduces as the temperature decreases with time. This procedure does not limit the new points to minimal points, this means that new points can have higher objective function value, helping to avoid the local minimum. We use the \textit{simulannealbnd} function from MATLAB. 
\item The Grid Search (GS) algorithm consists of testing a combination of values for all the hyperparameters. This is a naive method that needs $N^M$ evaluations where $M$ is the number of hyperparameters and $N$ is the number of values for each hyperparameter. The advantage of this method is that it can be easily parallelized, however, the method itself does not define a maximum number of evaluations. Therefore, we have split the search space based on the lower and upper bounds of $C$ and $\gamma$ to obtain the same number of evaluations allowed for other methods. We implemented the method in MATLAB. 
\item The Random Search (RS) algorithm starts by generating a set of points in the pre-defined search space. Subsequently, it gets the minimum among them and refines the search around it. It repeats this process until achieving the stop criterion. We implemented the method in MATLAB.
\end{itemize}

As one may see we do not compare the proposed method with derivative or evolutionary methods. In the framework of BBO, which includes the hyperparameter optimization problem, the derivative is unavailable, making the former unsuitable, while the later consists of global search heuristic methods (e.g. genetic algorithm and particle swarm optimization) in which the emphasis is on finding a decent global solution instead of finding an accurate local solution that provides a stopping criterion with some assurance of optimality.

%In the framework of black-box optimization, which includes the hyperparameter optimization problem, the derivative is unavailable, which makes no sense in compare the proposed approach with finite difference-based methods. Besides, we do not compare the proposed approach with evolutionary methods, such as genetic algorithms and particle swarm optimization, which are global search heuristic methods (the emphasis is on finding a decent global solution instead of finding an accurate local solution), that do not guarantee success, i.e., they do not employ mathematically proven stopping criteria, in contrary to derivative-free optimization that provide a stopping test with some assurance of optimality. 

We have evaluated the hinge-loss output value and the classification accuracy in the test set after 100 evaluations using a common configuration for all methods and the Ortho-MADS with the Nelder-Mead search. We used the functions \textit{fitcsvm} and \textit{fitcecoc} from the \textit{MATLAB Statistics and Machine Learning Toolbox} to train the SVM for the binary or multiclass cases respectively, and the function \textit{predict} to evaluate the learned model in the test sets. We defined the lower bound of the search space as $L=[0.01, 0.01]$ and its upper bound as $U=[100, 100]$. Considering that the starting point plays an important role in BBO optimization, we have compared the methods using six different initialization points $x_0=(C,\gamma)=\{(0.5, 0.5), (10, 10), (50, 50), (90, 90), (1, 90), (90, 1)\}$. We have used the pre-defined validation set when it was available on the hinge-loss function, and for the cases where there was not a pre-defined validation set, we have used the training set with a stratified 3-fold cross validation strategy. For the RS and the GS, there is no pre-defined initialization point, and we have set a linear search of 100 iterations, i.e., $C\in\{1,10\times1,10\times2,\dots,10\times9\}$ and $\gamma\in\{1,10\times1,10\times2,\dots,10\times9\}$. 

Tab.~\ref{tab:allmethods} shows the mean accuracy, standard deviation and maximum accuracy for all hyperparameter tuning methods. Both Ortho-MADS and BADS have shown to be more consistent to achieve a competitive mean accuracy for all datasets. The BO, SA, and RS present competitive results in ten datasets, and the GS shows competitive results in six datasets. Tab.~\ref{tab:fminallmethods} presents the mean loss $\mathcal{L}$, its standard deviation and the minimum loss $\mathcal{L}_{min}$. For most of the datasets, the behavior is similar to that presented in Tab.~\ref{tab:allmethods}. The Bayesian, SA, RS, and GS provided the lowest $\mathcal{L}_{min}$ for the Splice dataset. However this result does not necessarily translate into high accuracy because in this case, the function might be overfitting. 

\begin{center}
\begin{table*}[htpb!]
%\hfill{}
%\renewcommand{\arraystretch}{1}
\footnotesize
\centering
%{\setlength{\tabcolsep}{3pt}
%\resizebox{1\hsize}{!}{%
\begin{tabular}{ll|l|l|l|l|l}
\hline
& Ortho-MADS & Bayesian & SA & RS & GS & BADS \\ [3pt]
\hline
Astro & \underline{0.970$\pm$0.001 $|$ 0.971} & \underline{0.970$\pm$0.000 $|$ 0.970} & 0.969$\pm$0.000 $|$ 0.970 & 0.955$\pm$0.004 $|$ 0.959 & 0.967$\pm$0.001 $|$ 0.967 & \underline{0.970$\pm$0.001 $|$ 0.972} \\ [3pt]
Car & 0.715$\pm$0.013 $|$ 0.732 & 0.695$\pm$0.034 $|$ 0.732 & 0.687$\pm$0.052 $|$ 0.732 & 0.407$\pm$0.230 $|$ 0.707 & 0.715$\pm$0.040 $|$ 0.732 & \underline{0.724$\pm$0.030 $|$ 0.780} \\ [3pt]
DNA & 0.942$\pm$0.000 $|$ 0.942 & 0.942$\pm$0.000 $|$ 0.942 & 0.616$\pm$0.005 $|$ 0.624 & 0.943$\pm$0.002 $|$ 0.945 & 0.943$\pm$0.001 $|$ 0.945 & \underline{0.945$\pm$0.003 $|$ 0.949} \\ [3pt]
Letter & 0.943$\pm$0.030 $|$ 0.957 & 0.954$\pm$0.000 $|$ 0.955 & 0.935$\pm$0.045 $|$ 0.959 & 0.838$\pm$0.022 $|$ 0.859 & 0.943$\pm$0.000 $|$ 0.943 & \underline{0.955$\pm$0.002 $|$ 0.958} \\ [3pt]
Madelon & 0.587$\pm$0.016 $|$ 0.607 & 0.573$\pm$0.003 $|$ 0.577 & 0.547$\pm$0.024 $|$ 0.565 & \underline{0.589$\pm$0.011 $|$ 0.607 }& 0.573$\pm$0.001 $|$ 0.575 & 0.578$\pm$0.012 $|$ 0.603 \\ [3pt]
Pendigits & \underline{0.973$\pm$0.001 $|$ 0.973} & 0.968$\pm$0.001 $|$ 0.970 & 0.971$\pm$0.002 $|$ 0.975 & 0.931$\pm$0.012 $|$ 0.956 & 0.859$\pm$0.000 $|$ 0.859 & \underline{0.973$\pm$0.002 $|$ 0.976} \\ [3pt]
Protein & 0.690$\pm$0.000 $|$ 0.691 & 0.690$\pm$0.001 $|$ 0.691 & \underline{0.692$\pm$0.000 $|$ 0.693} & 0.678$\pm$0.008 $|$ 0.694 & 0.691$\pm$0.001 $|$ 0.692 & 0.685$\pm$0.004 $|$ 0.689 \\ [3pt]
Satimage & 0.912$\pm$0.001 $|$ 0.912 & \underline{0.915$\pm$0.001 $|$ 0.917} & \underline{0.915$\pm$0.004 $|$ 0.918} & 0.876$\pm$0.016 $|$ 0.908 & 0.705$\pm$0.020 $|$ 0.718 & 0.910$\pm$0.003 $|$ 0.915 \\ [3pt]
Shuttle & 0.905$\pm$0.000 $|$ 0.905 & 0.913$\pm$0.005 $|$ 0.917 & \underline{0.918$\pm$0.001 $|$ 0.918} & 0.885$\pm$0.018 $|$ 0.910 & 0.718$\pm$0.000 $|$ 0.718 & 0.907$\pm$0.007 $|$ 0.917 \\ [3pt]
Splice & \underline{0.897$\pm$0.001 $|$ 0.899} & 0.607$\pm$0.005 $|$ 0.611 & 0.614$\pm$0.010 $|$ 0.626 & 0.870$\pm$0.020 $|$ 0.899 & 0.582$\pm$0.000 $|$ 0.582 & \underline{0.897$\pm$0.002 $|$ 0.901} \\ [3pt]
Svmguide4 & \underline{0.846$\pm$0.010 $|$ 0.859} & 0.774$\pm$0.004 $|$ 0.780 & 0.774$\pm$0.057 $|$ 0.824 & 0.540$\pm$0.117 $|$ 0.696 & 0.712$\pm$0.007 $|$ 0.720 & \underline{0.846$\pm$0.006 $|$ 0.856} \\ [3pt]
USPS & \underline{0.943$\pm$0.001 $|$ 0.944} & \underline{0.943$\pm$0.001 $|$ 0.944} & \underline{0.943$\pm$0.001 $|$ 0.944} & 0.942$\pm$0.003 $|$ 0.945 & \underline{0.943$\pm$0.001 $|$ 0.945} & \underline{0.943$\pm$0.001 $|$ 0.944} \\ [3pt]
Vowels & 0.622$\pm$0.013 $|$ 0.641 & \underline{0.630$\pm$0.011 $|$ 0.641} & 0.587$\pm$0.016 $|$ 0.608 & 0.498$\pm$0.052 $|$ 0.589 & 0.591$\pm$0.003 $|$ 0.593 & 0.625$\pm$0.011 $|$ 0.641 \\ [3pt]
 \hline
\end{tabular}
%}%
%}
\caption{Mean accuracy, standard deviation and maximum accuracy for all hyperparameter optimization methods in 13 datasets. The best results are underlined.}
\label{tab:allmethods}
\end{table*}
\end{center}

\begin{center}
\begin{table*}[htpb!]
\centering
\footnotesize
%{\setlength{\tabcolsep}{3pt}
%\resizebox{1\hsize}{!}{%
\begin{tabular}{ll|l|l|l|l|l}
\hline
& Ortho-MADS & Bayesian & SA & RS & GS & BADS \\ [3pt]
\hline
Astro & 1.042$\pm$0.001 $|$ 1.041 & \underline{1.041$\pm$0.001 $|$ 1.039} & 1.042$\pm$0.001 $|$ 1.039 & 1.082$\pm$0.015 $|$ 1.067 & 1.045$\pm$0.000 $|$ 1.044 & 1.043$\pm$0.002 $|$ 1.041 \\ [3pt]
Car & 1.190$\pm$0.003 $|$ 1.186 & 1.194$\pm$0.005 $|$ 1.186 & 1.194$\pm$0.004 $|$ 1.190 & 1.206$\pm$0.012 $|$ 1.186 & \underline{1.188$\pm$0.001 $|$ 1.186} & 1.191$\pm$0.004 $|$ 1.184 \\ [3pt]
DNA & \underline{1.033$\pm$0.000 $|$ 1.033} & \underline{1.033$\pm$0.000 $|$ 1.033} & 1.046$\pm$0.002 $|$ 1.044 & 1.034$\pm$0.001 $|$ 1.034 & 1.044$\pm$0.001 $|$ 1.042 & 1.035$\pm$0.002 $|$ 1.033 \\ [3pt]
Letter & \underline{1.001$\pm$0.000 $|$ 1.001} & \underline{1.001$\pm$0.000 $|$ 1.001} & 1.006$\pm$0.007 $|$ 1.003 & 1.003$\pm$0.001 $|$ 1.002 & \underline{1.001$\pm$0.000 $|$ 1.001} & \underline{1.001$\pm$0.000 $|$ 1.001} \\ [3pt]
Madelon & 1.459$\pm$0.006 $|$ 1.453 & 1.450$\pm$0.004 $|$ 1.445 & 1.455$\pm$0.005 $|$ 1.446 & 1.498$\pm$0.024 $|$ 1.462 & \underline{1.449$\pm$0.003 $|$ 1.444} & 1.452$\pm$0.009 $|$ 1.443 \\ [3pt]
Pendigits & \underline{1.001$\pm$0.000 $|$ 1.001} & 1.041$\pm$0.001 $|$ 1.040 & 1.042$\pm$0.002 $|$ 1.040 & 1.090$\pm$0.012 $|$ 1.068 & 1.043$\pm$0.001 $|$ 1.042 & \underline{1.001$\pm$0.000 $|$ 1.001} \\ [3pt]
Protein & \underline{1.157$\pm$0.000 $|$ 1.157} & \underline{1.157$\pm$0.000 $|$ 1.157} & 1.163$\pm$0.000 $|$ 1.163 & 1.168$\pm$0.011 $|$ 1.158 & 1.163$\pm$0.000 $|$ 1.162 & 1.158$\pm$0.000 $|$ 1.157 \\ [3pt]
Satimage & \underline{1.011$\pm$0.000 $|$ 1.011} & 1.041$\pm$0.001 $|$ 1.039 & 1.042$\pm$0.001 $|$ 1.040 & 1.090$\pm$0.020 $|$ 1.056 & 1.045$\pm$0.001 $|$ 1.043 & \underline{1.011$\pm$0.000 $|$ 1.011} \\ [3pt]
Shuttle & \underline{1.000$\pm$0.000 $|$ 1.000} & 1.042$\pm$0.001 $|$ 1.041 & 1.042$\pm$0.000 $|$ 1.041 & 1.080$\pm$0.025 $|$ 1.047 & 1.044$\pm$0.001 $|$ 1.043 & \underline{1.000$\pm$0.000 $|$ 1.000} \\ [3pt]
Splice & 1.179$\pm$0.005 $|$ 1.175 & \underline{1.041$\pm$0.001 $|$ 1.040} & 1.042$\pm$0.001 $|$ 1.041 & 1.076$\pm$0.012 $|$ 1.061 & 1.044$\pm$0.001 $|$ 1.043 & 1.185$\pm$0.004 $|$ 1.178 \\ [3pt]
SVMguide4 & 1.044$\pm$0.002 $|$ 1.041 & \underline{1.041$\pm$0.001 $|$ 1.040} & 1.045$\pm$0.004 $|$ 1.041 & 1.089$\pm$0.013 $|$ 1.072 & 1.045$\pm$0.001 $|$ 1.044 & 1.044$\pm$0.001 $|$ 1.042 \\ [3pt]
USPS & \underline{1.002$\pm$0.000 $|$ 1.002} & \underline{1.002$\pm$0.000 $|$ 1.002} & \underline{1.002$\pm$0.000 $|$ 1.002} & 1.003$\pm$0.001 $|$ 1.002 & \underline{1.002$\pm$0.000 $|$ 1.002} & \underline{1.002$\pm$0.000 $|$ 1.002} \\ [3pt]
Vowels & \underline{1.003$\pm$0.000 $|$ 1.003} & \underline{1.003$\pm$0.000 $|$ 1.003} & 1.194$\pm$0.004 $|$ 1.188 & 1.023$\pm$0.010 $|$ 1.006 & 1.004$\pm$0.000 $|$ 1.004 & 1.004$\pm$0.000 $|$ 1.003 \\ [3pt]
 \hline
\end{tabular} 
% }%
% }
\caption{Mean loss ($\mathcal{L}$), its standard deviation and minimum loss ($\mathcal{L}_{min}$) for all hyperparameter optimization methods in 13 datasets. The best results are underlined.}
\label{tab:fminallmethods}
\end{table*}
\end{center}

\begin{figure*}[htpb!]
\subcaptionbox{\label{fig:sat_conv}}{\includegraphics[width=0.49\textwidth]{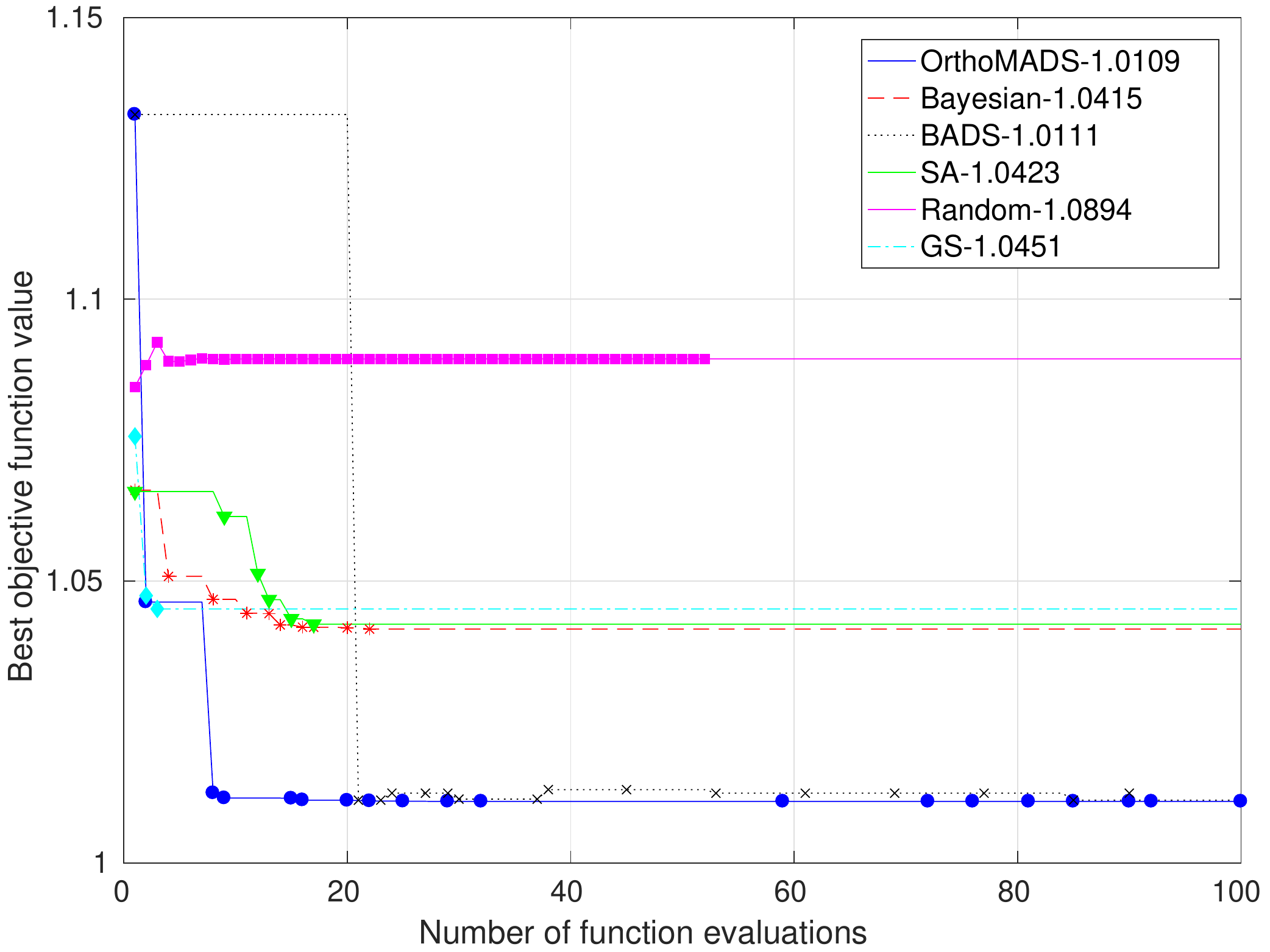}}
\hfill
\subcaptionbox{\label{fig:sat_traj}}{\includegraphics[width=0.49\textwidth]{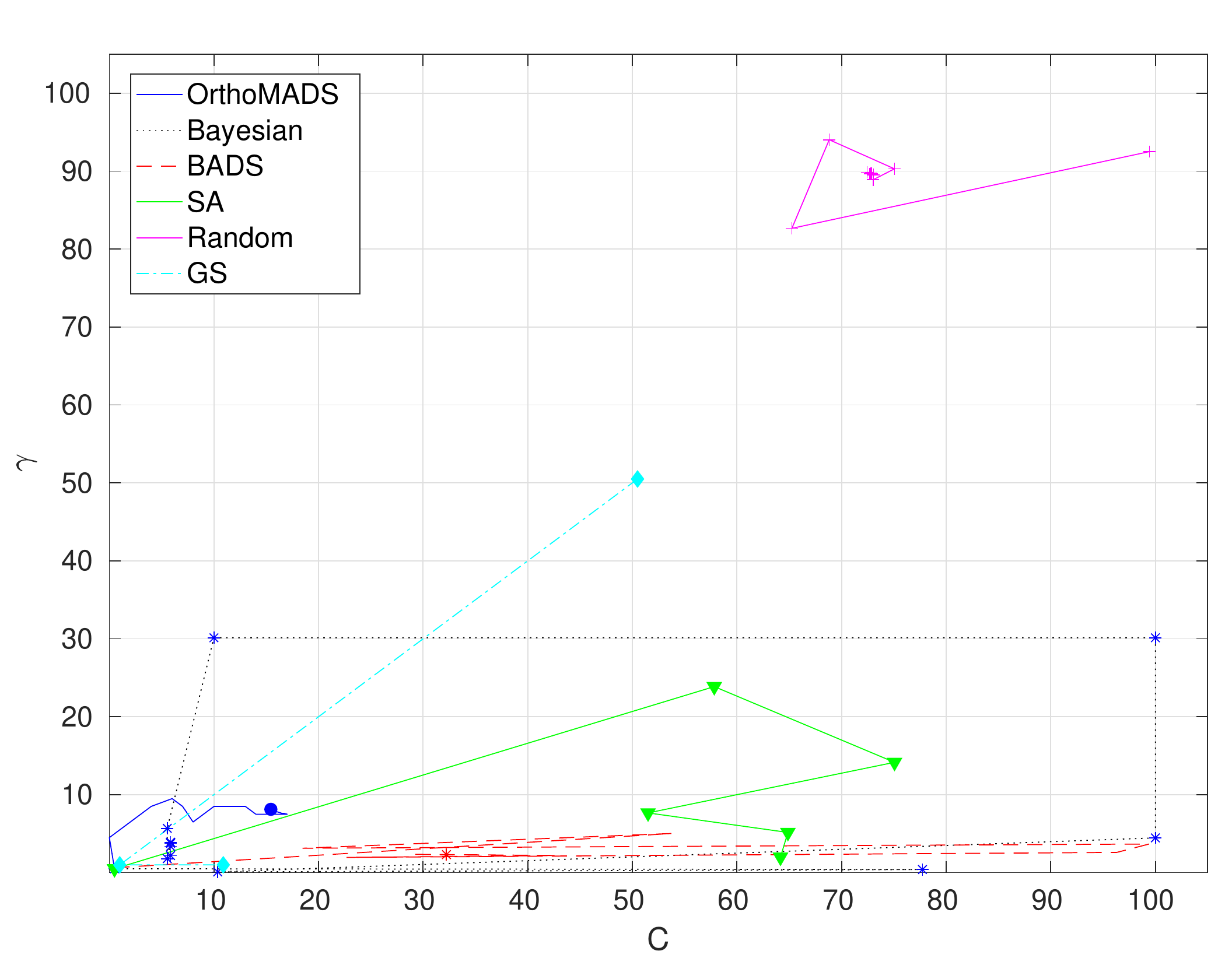}}
\label{fig:satimage}
\caption{Satimage dataset with $x_0=\{0.5,0.5\}$: (a) convergence plot comparison; (b) trajectory plot comparison.}
\end{figure*}

\begin{figure*}[htpb!]
\subcaptionbox{\label{fig:ast005}}{\includegraphics[width=0.49\textwidth]{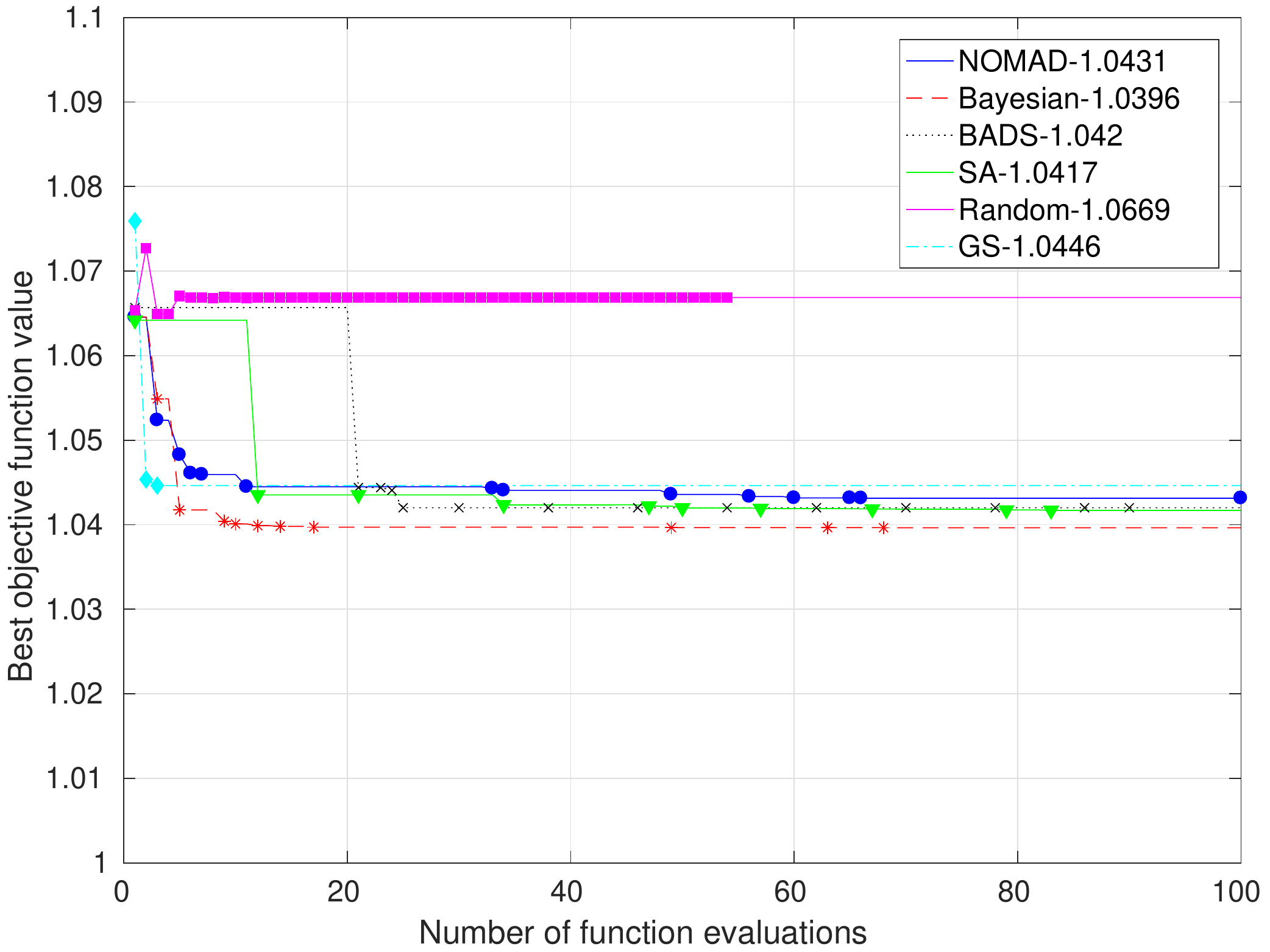}}
\hfill
\subcaptionbox{\label{fig:ast901}}{\includegraphics[width=0.49\textwidth]{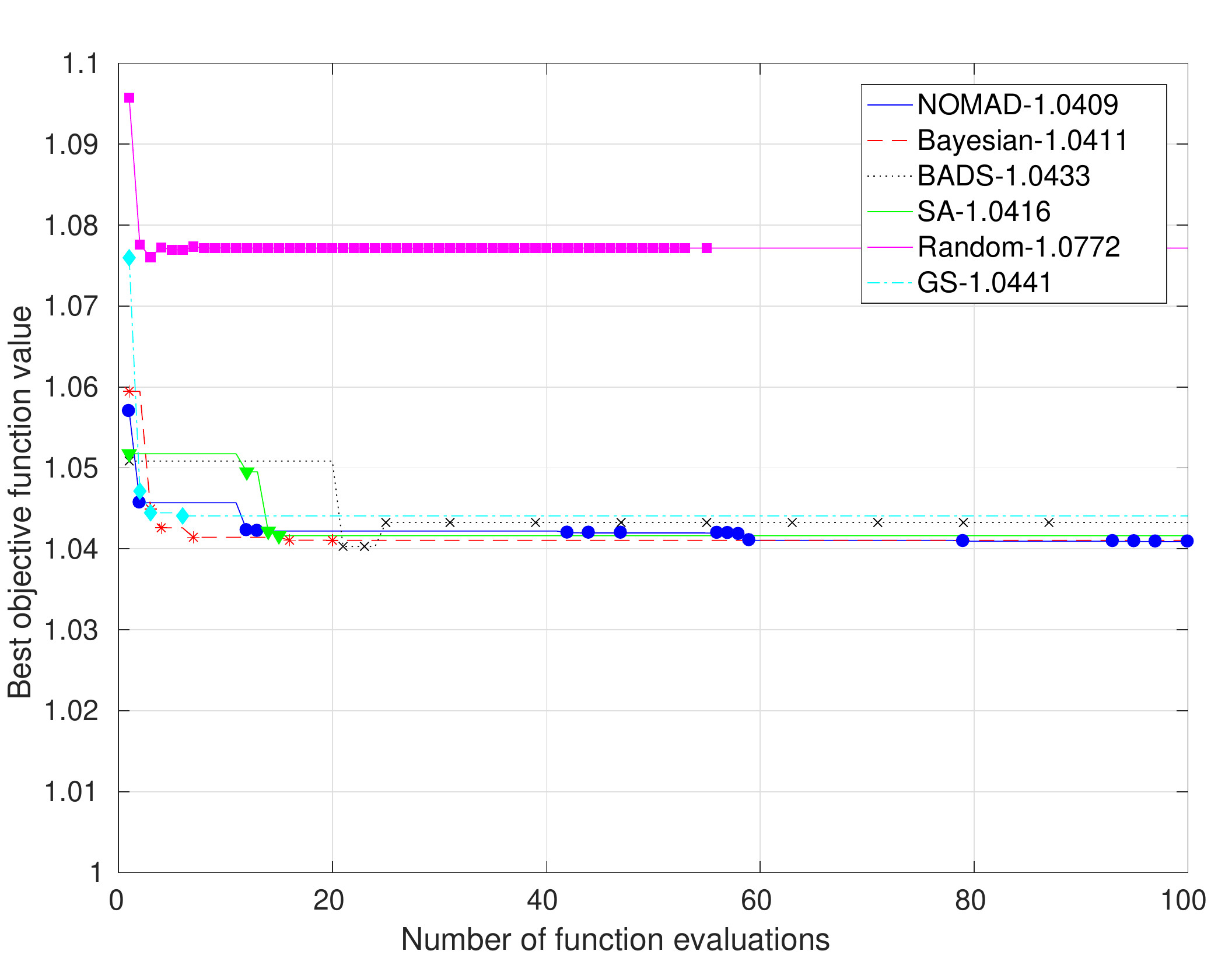}}
\label{fig:astro}
\caption{Astroparticle dataset comparison convergence plot with different starting points: (a) at $x_0=\{0.5,0.5\}$; (b) at $x_0=\{90,1\}$.}
\end{figure*}

Another aspect to analyze is the convergence rate and the trajectory of the algorithms. For all datasets, the Ortho-MADS presents a competitive convergence rate, and in many cases (as exemplified in Figs.~\ref{fig:sat_conv} and \ref{fig:sat_traj}), both the Ortho-MADS and the BADS (that also uses the MADS algorithm) have the fastest convergence rate to reach a minimum. In some cases, the Ortho-MADS may not have the fastest convergence, as depicted in Figs.~\ref{fig:ast005} and \ref{fig:ast901}. However, it is still competitive with other methods, and may pass over its convergence rate to reach a lower local minimum, as shown in Fig.~\ref{fig:ast901}. The BADS achieved the best results overall, with better accuracy in eight out of thirteen datasets (Astro, Car, DNA, Letter, Pendigits, Splice, Svmguide4, and USPS), and competitive accuracy for all other datasets with a low standard deviation. The Ortho-MADS has the second-best results, with best results in five out of thirteen datasets (Astro, Pendigits, Splice, Svmguide4, and USPS), and competitive accuracy for all other datasets with a low standard deviation. 

The Bayesian and the SA methods presented the best results in four out of thirteen datasets, but sometimes the results are not competitive, as observed for the Splice and Svmguide4 datasets when applying the Bayesian method, and for the DNA, Splice, and Svmguide4 datasets when using the SA method. In addition, both methods have presented standard deviations higher than Ortho-MADS and BADS. The RS achieved the best accuracy in the Madelon dataset, however, RS depends on the randomness that leads to more iterations to achieve a good result, and GS depends on the grid choice, which creates unreachable spaces.

A close look at the Ortho-MADS standard deviation (this extends to other methods as well) from Tab.~\ref{tab:allmethods} indicates that in some cases we do not reach the best point, and this fact could be related to the choice of the starting point, as it has an important influence on the result or the method randomness that falls into a local minimum. Figs.~\ref{fig:ast005} and \ref{fig:ast901} exemplify the influence of the starting point on the effectiveness of the algorithms. From the starting point $x_0=\{0.5,0.5\}$, depicted in Fig.~\ref{fig:ast005}, the Ortho-MADS, Bayesian, BADS, SA, and GS achieved worse objective function value when compared to the starting point $x_0=\{90,1\}$ from Fig.~\ref{fig:ast901}. 

We generate an ordering of the methods based on the mean, maximum and worst accuracy reported in Tab.~\ref{tab:allmethods}. Tab.~\ref{tab:avgrank} summarizes the comparison between all methods through an average ranking \cite{Brazdil2000} according to the measured accuracy mean, worst case, and best case. The BADS has the best rank among all considered methods, followed by the Ortho-MADS, Bayesian, SA, GS, and RS. The most consistent methods are the BADS and the Ortho-MADS, ranking 1 and 2 for the best mean and the best maximum accuracy, and 6 and 5 for the worst mean accuracy respectively.

\begin{table}[htpb!]
\footnotesize
\centering
\begin{tabular}{l|c|c|c|c|c|c}
\hline
 & \multicolumn{2}{c|}{Best Mean} & \multicolumn{2}{c|}{Worst Mean} & \multicolumn{2}{c}{Best Maximum} \\
Algorithm & \multicolumn{2}{c|}{Accuracy} & \multicolumn{2}{c|}{Accuracy} & \multicolumn{2}{c}{Accuracy} \\ \cline{2-7}
& $\bar{r}$ & Rank & $\bar{r}$ & Rank & $\bar{r}$ & Rank \\
\hline
BADS & 1.77 & 1 & 4.38 & 6 & 2.08 & 1\\ %8 + 2 + 9 + 4 = 23 / 13
Ortho-MADS & 2.07 & 2 & 3.77 & 5 & 2.69 & 2\\ % 5 + 6 + 12 + 4 = 27 / 13
Bayesian & 2.31 & 3 & 3.46 & 4 & 3.23 & 4\\ %4 + 6 + 12 + 8 = 30 / 13
SA & 2.78 & 4 & 3.08 & 3 & 2.92 & 3\\ %4 + 6 + 3 + 12 + 5 + 6 = 36 / 13
GS & 3.54 & 5 & 2.38 & 2 & 4.38 & 6\\ %1 + 6 + 9 + 4 + 20 + 6 = 46 / 13
RS & 3.77 & 6 & 2.08 & 1 & 3.92 & 5\\ %1 + 6 + 0 + 16 + 20 + 6 = 49 / 13
\hline
\end{tabular}
\caption{Average ranking (AR) considering the best mean accuracy, the worst mean accuracy and the maximum accuracy for the 13 datasets \cite{Brazdil2000}.}
\label{tab:avgrank}
\end{table}

\subsection{Proposed Approach}
%%%%%%%%%%%%%%%%%%%%%%%%%%%%%%%%%%%%%%%%%%

%Rewrite

The Ortho-MADS has shown to be competitive with other state-of-the-art methods (Tab.~\ref{tab:allmethods}) to tune the hyperparameters of the SVM with Gaussian kernel, however, we propose to combine the Ortho-MADS convergence properties with two different search algorithms to enhance the stability and reachability, and to use the mesh size as stopping criterion. The NM search strategy leads to a faster convergence when compared to regular Ortho-MADS search strategy, i.e., it requires fewer function evaluations to reach the pre-defined minimum mesh size. The VNS explores regions far from the incumbent (increasing the number of function evaluations), helping to escape from an eventual undesired local minimum. The VNS counterbalance the NM fast convergence; however, it explores more regions from the search space and it mitigates the initial point influence. The Ortho-MADS attempts to find an accurate local solution and using the mesh-size as stopping criterion translates into stopping the algorithm when achieving a local solution (mesh-size) that satisfies the user needs. 

We evaluate the accuracy and the number of function evaluations for the Ortho-MADS with two search algorithms (NM and VNS, both with non-opportunistic strategy\footnote{For each search iteration, the algorithm does not finish when a better incumbent is found. It only finishes when all points are evaluated.}), using the minimum mesh size as stopping criterion. Because the MADS direction is randomly chosen, for each dataset we run 50 times using the following default configuration (empirically defined): the lower bound $(L)$ as $[0.01, 0.01]$, the upper bound $(U)$ as $[100.01, 100.01]$, the starting point $(x_0)$ as $\{50, 50\}$, the minimum mesh size $(\delta_{min})$ as $0.009$ (that corresponds to three shrinking executions from the initial mesh size), and the perturbation amplitude is $\xi=0.25$. From the initial configuration, we further analyze the impact of changing the starting point $x_0$, the minimum mesh size $\delta_{min}$ and the perturbation amplitude $\xi$. Tab.~\ref{tab:all_VNS} presents the mean accuracy, the standard deviation, and the maximum and median accuracy. For each measure we also present the corresponding number of function evaluations. As stated before, the incorporation of VNS in the search step aids the Ortho-MADS to escape from local minimum, which may lead to better results, and the NM counterbalance the number of function evaluations needed.

\begin{table}[htpb!]
\footnotesize
\centering
%\resizebox{0.99\hsize}{!}{
\begin{tabular}{lllll}\hline
 & Mean & Std & Max & Median \\\hline
Astro & 0.968 $|$ 167 & 0.001 $|$ 45 & 0.971 $|$ 81 & 0.969 $|$ 162 \\ 
Car$^\star$ & 0.720 $|$ 141 & 0.027 $|$ 41 & 0.829 $|$ 105 & 0.707 $|$ 137 \\ 
DNA & 0.942 $|$ 51 & 0.000 $|$ 0 & 0.942 $|$ 51 & 0.942 $|$ 51 \\
Letter$^\star$ & 0.954 $|$ 111 & 0.000 $|$ 0 & 0.954 $|$ 111 & 0.954 $|$ 111 \\
Madelon$^\star$ & 0.598 $|$ 85 & 0.007 $|$ 43 & 0.607 $|$ 70 & 0.602 $|$ 75 \\
Pendigits & 0.972 $|$ 117 & 0.000 $|$ 27 & 0.973 $|$ 76 & 0.972 $|$ 113 \\
Protein & 0.691 $|$ 73 & 0.000 $|$ 0 & 0.691 $|$ 73 & 0.691 $|$ 73 \\
Satimage & 0.913 $|$ 108 & 0.000 $|$ 0 & 0.913 $|$ 108 & 0.913 $|$ 108 \\
Shuttle$^\star$ & 0.999 $|$ 73 & 0.000 $|$ 0 & 0.999 $|$ 73 & 0.999 $|$ 73 \\
Splice & 0.897 $|$ 120 & 0.001 $|$ 27 & 0.901 $|$ 78 & 0.897 $|$ 118 \\
Svmguide4$^\star$ & 0.847 $|$ 118 & 0.008 $|$ 23 & 0.865 $|$ 135 & 0.848 $|$ 112 \\
USPS & 0.943 $|$ 95 & 0.001 $|$ 23 & 0.945 $|$ 78 & 0.943 $|$ 97 \\
Vowels & 0.624 $|$ 128 & 0.011 $|$ 32 & 0.643 $|$ 91 & 0.621 $|$ 123 \\\hline
\end{tabular}
%}
\caption{Accuracy $|$ No.~of function evaluations for the proposed approach. Mean, standard deviation, maximum (and the number of function evaluations regarding the best result), and median for all 13 datasets. $^\star$ indicates higher mean accuracy. }
\label{tab:all_VNS}
\end{table}

Tab.~\ref{tab:all_VNS} presents the results of the proposed approach, and comparing with Tab.~\ref{tab:allmethods} we observe several improvements. Using the minimum mesh-size as stopping criterion may avoid unnecessary function evaluations. In our previous experiment (Tab.~\ref{tab:allmethods}), the stopping criterion was 100 function evaluations, and using the new stopping criterion we reach the same or better accuracy with fewer function evaluations (as reported in Tab.~\ref{tab:all_VNS}) in all runs for five datasets (DNA, Madelon, Protein, Shuttle, and USPS). In addition, it may reduce the number of function evaluations in another five datasets (Astro, Pendigits, Shuttle, Splice, and Vowels), i.e., sometimes it achieves the minimum mesh-size with fewer than 100 function evaluations. Regarding the stability, the DNA, Letter, Protein, Satimage, and Shuttle datasets present a standard deviation of approximately zero. We achieved a better accuracy for the Car dataset (from 0.780 to 0.829), but it increases the number of function evaluations to reach the minimum mesh-size. For the Madelon dataset we reached the best accuracy reported in Tab.~\ref{tab:allmethods} by the RS algorithm, and increased the mean accuracy with fewer function evaluations (85 of mean, and best value achieved with 70). We achieved the best accuracy overall for the Shuttle dataset, with a lower number of function evaluations. In the Vowels dataset, the best accuracy improved from 0.641 to 0.643, however, with an increase in the number of function evaluations (from 100 to 128 of mean). 

Here again, we generate an ordering of the methods but now replacing the Ortho-Mads by the proposed approach. Tab.~\ref{tab:avgrank2} summarizes the comparison between all methods through an average ranking \cite{Brazdil2000} according to the measured accuracy mean, worst case, and best case. The proposed approach has the best rank among all considered methods, followed by the BADS, Bayesian, SA, GS, and RS. The most consistent methods are the proposed approach and the BADS, ranking 1 and 2 for the best mean and the best maximum accuracy, and 5 and 6 for the worst mean accuracy respectively. 

\begin{table}[htpb!]
\footnotesize
\centering
%\begin{tabular}{l|c|c|c|c|c|c}
%\hline
%Algorithm & \multicolumn{2}{c}{AR} \\ \cline{2-3}
%& $\bar{r}$ & Rank \\
%\hline
\begin{tabular}{l|c|c|c|c|c|c}
\hline
 & \multicolumn{2}{c|}{Best Mean} & \multicolumn{2}{c|}{Worst Mean} & \multicolumn{2}{c}{Best Maximum} \\
Algorithm & \multicolumn{2}{c|}{Accuracy} & \multicolumn{2}{c|}{Accuracy} & \multicolumn{2}{c}{Accuracy} \\ \cline{2-7}
& $\bar{r}$ & Rank & $\bar{r}$ & Rank & $\bar{r}$ & Rank \\
\hline
%Ortho-MADS (NM+VNS)& 1.85 & 1 & 4.31 & 5 & 2.15 & 1\\ 
\bf{Proposed Approach} & 1.85 & 1 & 4.31 & 5 & 2.15 & 1\\ 
BADS & 1.92 & 2 & 4.46 & 6 & 2.38 & 2\\
Bayesian & 2.61 & 3 & 3.38 & 4 & 3.46 & 4\\
SA & 3.08 & 4 & 3.08 & 3 & 3.15 & 3\\
GS & 3.85 & 5 & 2.31 & 2 & 4.46 & 6\\
RS & 4.15 & 6 & 2.00 & 1 & 4.08 & 5\\
\hline
\end{tabular}
\caption{Average ranking (AR) considering the best mean accuracy, the worst mean accuracy and the maximum accuracy for the 13 datasets \cite{Brazdil2000}.}
\label{tab:avgrank2}
\end{table}

\begin{figure}[htpb!]
\centering
\includegraphics[width=0.45\textwidth]{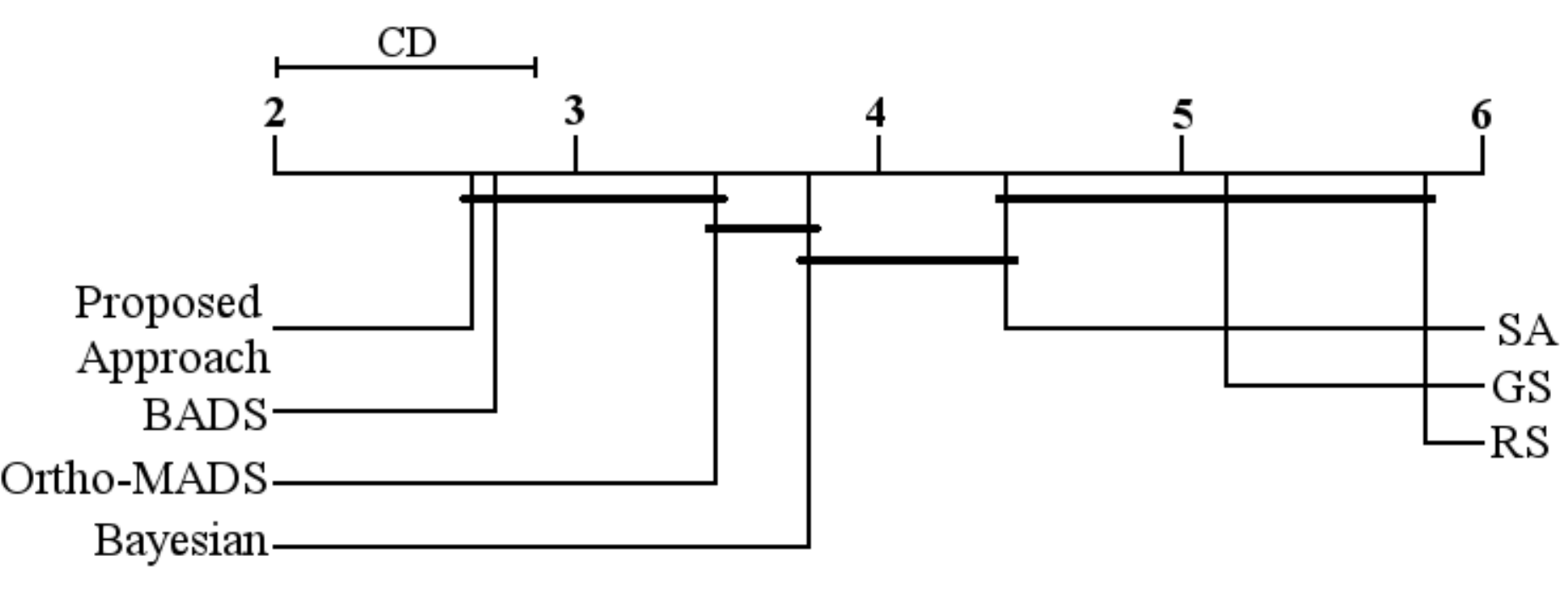}
\caption{Nemenyi test critical difference.}
\label{fig:nemenyi_h0}
\end{figure}

The Friedman rank sum test shows a p-value of 0.00024, and Tab.~\ref{tab:nemenyi} presents the Nemenyi test using Tabs.~\ref{tab:allmethods} and \ref{tab:all_VNS}. The results indicate that the proposed approach (Ortho-MADS+VNS+NM) presents similar results to BADS, and the concordance of results between Ortho-MADS and BADS are smaller than the proposed approach. We can conclude that both BADS and Ortho-MADS+VNS+NM present superior performance compared to other methods. Furthermore, the advantage of the proposed approach when compared to BADS is the false minimum avoidance and the stopping criterion. Fig. \ref{fig:nemenyi_h0} depicts the critical difference graph comparing all methods, illustrating the results of Tab. \ref{tab:nemenyi}. The confidence interval is 95\% for the null hypotheses of $H_0: \Theta_i = \Theta_j$ and the alternative hypotheses of $H_1: \Theta_i \neq \Theta_j$. Considering the used confidence interval the p-values close to one do not reject the null hypotheses, while the p-values close to zero reject the $H_0$ and do not reject $H_1$.

\begin{table}[htbp!]
\centering
\footnotesize
{\setlength{\tabcolsep}{3pt}
\resizebox{1\hsize}{!}{
\begin{tabular}{l|cccccc}
\hline
& Proposed  & Ortho-MADS & Bayesian & SA & RS & GS \\
& Approach & &  &  &  &  \\
\hline
Ortho-MADS & 0.9637 & - & - & - & - & - \\
Bayesian & 0.8446 & 0.9998 & - & - & - & - \\
SA & 0.3597 & 0.9175 & 0.9876 & - & - & - \\
RS & 0.0037 & 0.0821 & 0.1959 & 0.6601 & - & - \\
GS & 0.0497 & 0.4163 & 0.6601 & 0.9780 & 0.9876 & - \\
BADS & 1.0000 & 0.9780 & 0.8844 & 0.4163 & 0.0052 & 0.0642 \\
\hline
\end{tabular}
}}
\caption{Nemenyi test between all methods using the 13 datasets.}
\label{tab:nemenyi}
\end{table}

We choose the Car dataset (as it has the largest standard deviation among all datasets reported in Tab.~\ref{tab:all_VNS}) to analyze the impact of changing the VNS, the starting point and the minimum mesh size. We start considering different perturbation amplitudes $\xi=\{0.25, 0.5,$ $0.75, 0.9\}$, and Tab.~\ref{tab:car_VNS} shows the results after 50 runs for each $\xi$. By increasing $\xi$, the algorithm can reach regions more distant from the current incumbent, however, it requires more function evaluations to achieve a high mean accuracy. A large $\xi$ does not translate into a better accuracy, as increasing $\xi$ it creates sparser points to evaluate which may skip good regions, as the results for $\xi=0.75$ and $\xi=0.9$ indicate.

\begin{table}[htpb!]
\footnotesize
\centering
%\resizebox{0.99\hsize}{!}{
\begin{tabular}{lllll} \hline
Car & $\xi=0.25$ & $\xi=0.5$ & $\xi=0.75$ & $\xi=0.9$ \\ \hline
Mean & 0.720 $|$ 141 & 0.720 $|$ 172 & 0.710 $|$ 275 & 0.710 $|$ 362 \\ 
Std & 0.027 $|$ 41.3 & 0.042 $|$ 60.5 & 0.042 $|$ 79.4 & 0.029 $|$ 86.9 \\ 
Max & 0.829 $|$ 105 & 0.829 $|$ 107 & 0.805 $|$ 200 & 0.780 $|$ 205 \\ 
Median & 0.707 $|$ 137 & 0.707 $|$ 161 & 0.707 $|$ 264 & 0.707 $|$ 361\\ \hline
\end{tabular} 
%}
\caption{Mean accuracy, standard deviation, maximum accuracy and median accuracy and the corresponding No.~of function evaluations by changing the VNS for the Car dataset for different perturbation amplitudes.}
\label{tab:car_VNS}
\end{table}

Tab.~\ref{tab:car_VNS2} shows that by decreasing $\delta_{min}$, the proposed approach needs more function evaluations to reach the desired local minimum, as the mesh size reduction occurs in sequence (it is not possible to execute two mesh reduction operations in the same Ortho-MADS with NM and VNS iteration). For $\delta_{min(C,\gamma)}=9e-1$ we need fewer function evaluations to reach the stopping criterion, however, the maximum accuracy achieved was lower than the results from $\delta_{min(C,\gamma)}=9e-3$. A smaller minimum mesh size, $\delta_{min(C,\gamma)}=9e-7$, does not guarantee a good performance, but for sure it increases the number of function evaluations needed to reach the stopping criterion. In this case the model can discard good points or over-fit the model. 

\begin{table}[htpb!]
\footnotesize
\centering
%\resizebox{0.99\hsize}{!}{
\begin{tabular}{llll} \hline
Car & $\delta_{min(C,\gamma)}=9e-1$  & $\delta_{min(C,\gamma)}=9e-3$  & $\delta_{min(C,\gamma)}=9e-7$ \\ \hline
Mean & 0.738 $|$ 34.9 & 0.729 $|$ 184.4  & 0.716 $|$ 236.7 \\
Std & 0.036 $|$ 21.3 & 0.035 $|$ 43.9& 0.029 $|$ 68.2 \\
Max & 0.804 $|$ 60 & 0.829 $|$ 133 & 0.804 $|$ 212 \\
Median & 0.756 $|$ 24 & 0.732 $|$ 169 & 0.707 $|$ 218 \\ \hline
\end{tabular}
%}
\caption{Mean accuracy, standard deviation, maximum accuracy and median accuracy and the corresponding No.~of function evaluations for the Car dataset for different minimum mesh sizes ($\delta_{min}$).}
\label{tab:car_VNS2}
\end{table}

We have also evaluated the influence of the starting point for the Car dataset using five pre-defined and five random starting points $x_0=\{(0.5, 0.5), (50, 50), (90, 90),$ $(1, 90), (90, 1), (70.93, 75.21), (50.60, 64.29), (25.21, 79.05),$ $(89.59, 13.49), (2.37, 57.91)\}$.

\begin{figure}[htpb!]
\centering
\includegraphics[width=0.45\textwidth]{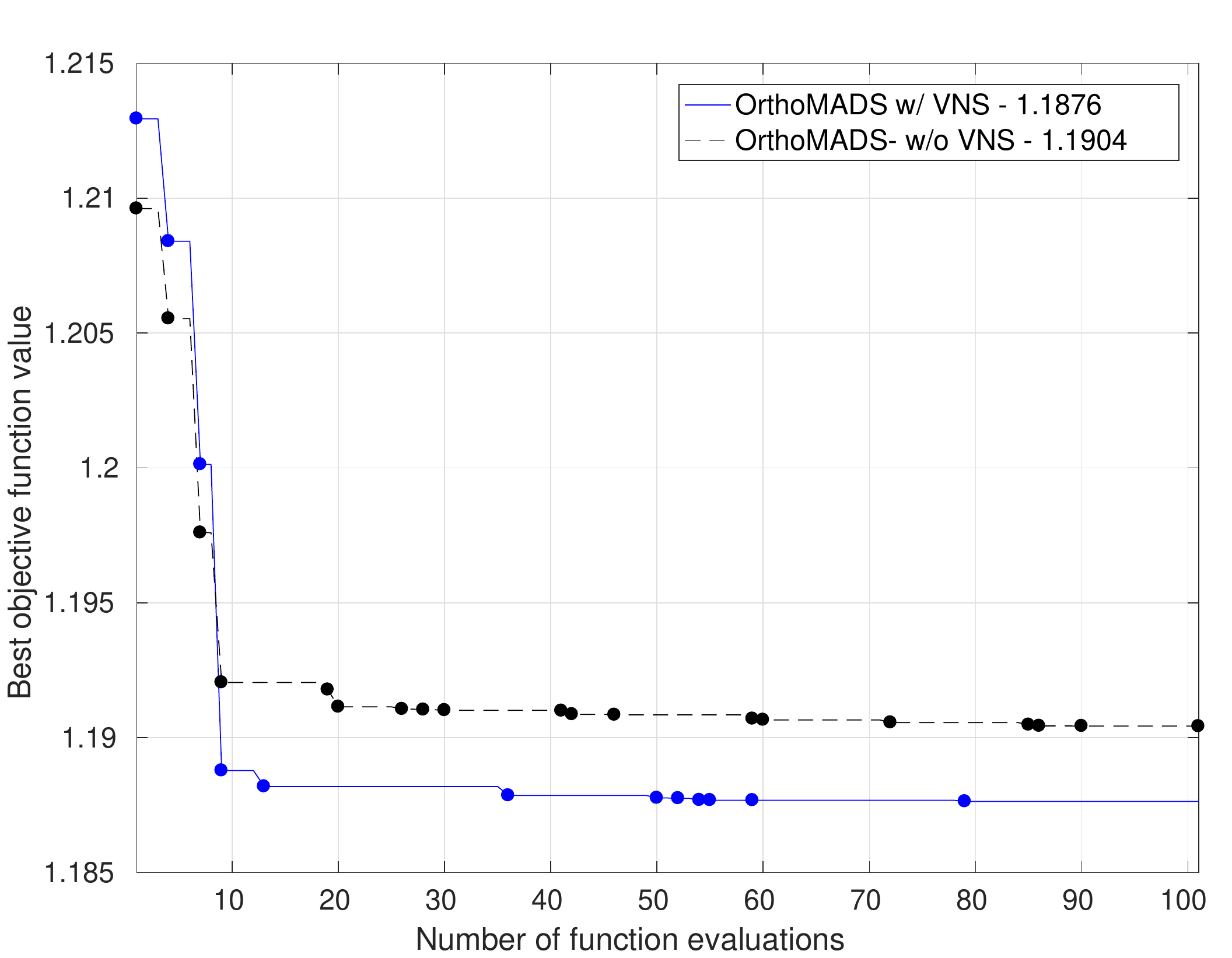}
\caption{Convergence plot comparison between Ortho-MADS with and without VNS for the Car dataset.}
\label{fig:NOMAD_Car}
\end{figure}

Tab.~\ref{tab:car_x0} shows that the mean value may be similar to the solution without VNS (as shown in Tab.~\ref{tab:allmethods}), however, we could reach at least $0.805$ of accuracy for all starting points, which is higher than the best solution from Tab.~\ref{tab:allmethods} ($0.780$ using BADS). Thus, the VNS mitigates the starting point effect and the MADS randomness. Fig.~\ref{fig:NOMAD_Car} exemplifies a comparison example between the Ortho-MADS with and without the VNS, both using the same starting point at $x_0=(50, 50)$ and the number of function evaluations. The Ortho-MADS without VNS reached $\mathcal{L}_{min}=1.1904$ and the accuracy of 0.7073 on the test set, while the Ortho-MADS with VNS reached $\mathcal{L}_{min}=1.1876$ and the accuracy of 0.8048. 

\begin{table}[htpb!]
\footnotesize
\centering
%\resizebox{0.99\hsize}{!}{
\begin{tabular}{llll} \hline
$x_0=(C,\gamma)$ & Mean & Max & Min \\ \hline 
$(0.5, 0.5)$ & 0.717 $|$ 1.191 & 0.805 $|$ 1.187 & 0.780 $|$ 1.185 \\
$(50, 50)$ & 0.735 $|$ 1.189 & 0.829 $|$ 1.187 & 0.805 $|$ 1.183 \\
$(90, 90)$ & 0.706 $|$ 1.190 & 0.805 $|$ 1.188 & 0.732 $|$ 1.183 \\
$(1, 90)$ & 0.690 $|$ 1.192 & 0.805 $|$ 1.194 & 0.780 $|$ 1.185 \\
$(90, 1)$ & 0.721 $|$ 1.192 & 0.805 $|$ 1.196 & 0.707 $|$ 1.118 \\
$(70.93, 75.21)$ & 0.698 $|$ 1.192 & 0.829 $|$ 1.188 & 0.707 $|$ 1.186 \\
$(50.60, 64.29)$ & 0.728 $|$ 1.191 & 0.805 $|$ 1.196 & 0.780 $|$ 1.184 \\
$(25.21, 79.05)$ & 0.726 $|$ 1.190 & 0.805 $|$ 1.189 & 0.707 $|$ 1.187 \\
$(89.59, 13.49)$ & 0.727 $|$ 1.191 & 0.805 $|$ 1.193 & 0.732 $|$ 1.185 \\
$(2.37, 57.91)$ & 0.727 $|$ 1.190 & 0.805 $|$ 1.189 & 0.780 $|$ 1.186 \\ \hline
\end{tabular} 
%}
\caption{Mean accuracy, maximum accuracy and minimum accuracy and the corresponding $\mathcal{L}$ value for the Car dataset for different starting points $x_0$.}
 \label{tab:car_x0}
\end{table}

Tab.~\ref{tab:madelon_it} compares the number of function evaluations that each method takes to achieve its best accuracy in the Madelon dataset. In the case of a good starting point (as we previously knew), all methods need the same or fewer function evaluations than the proposed approach to achieve their best accuracy. However, none can reach the accuracy achieved by the proposed method. In the case of a "bad" starting point, the BADS, Bayesian, and SA have a high probability of achieving an undesired local minimum, not reaching the highest accuracy. The GS and the RS depend on the grid designed by the user and on the dataset randomness to find a point that results in the best accuracy. Tab.~\ref{tab:car_it} shows the accuracy of each method using the accuracy of 0.8 or one thousand function evaluations as stopping criteria and previously known "good" starting points. We choose the Car dataset because it has a significant difference in the best accuracy between the proposed approach and the other methods, and no other method achieved accuracy above 0.732.

We noticed in experiments that all methods were susceptible to fall at a minimum point that is not their best. Even the Ortho-MADS method without the VNS approach can reach this condition. The VNS approach is important to leave false minimum points. Ortho-MADS+VNS is less sensitive to the impact of initialization points and the false minimum problem. During the experiments we observed one situation where the RS achieved its minimum at five iterations with a random initialization point, but in the great majority of the situations, this method could not reproduce this result with less than 1,000 iterations. 

\begin{table}[htpb!]
\footnotesize
\centering
%\resizebox{0.99\hsize}{!}{
\begin{tabular}{lll}
\hline
Approach & Accuracy & \# Evaluations \\
 \hline
\bf{Proposed Approach} & 0.607 &  70  \\
Bayesian & 0.568 & 30  \\
BADS & 0.602 & 106  \\
GS & 0.571 & 57  \\
RS & 0.603 & 30  \\
SA & 0.601 & 19  \\
 \hline
\end{tabular} 
%}
\caption{Best maximum accuracy and No.~of function evaluations for the Madelon dataset.}
 \label{tab:madelon_it}
\end{table}

\begin{table}[htpb!]
\footnotesize
\centering
%\resizebox{0.99\hsize}{!}{
\begin{tabular}{lll}
\hline
Approach & Accuracy & \# Evaluations\\
 \hline
\bf{Proposed Approach} & 0.829 & 105 \\
Bayesian & 0.707 & 1,000 \\
BADS &  0.707 & 1,000 \\
GS &  0.732 & 1,000 \\
RS &  0.707 & 1,000 \\
SA &  0.732 & 1,000 \\
 \hline
\end{tabular} 
%}
\caption{Best maximum accuracy and No.~of function evaluations for the Car dataset with stopping condition of either 0.8 of accuracy or 1,000 function evaluations.}
 \label{tab:car_it}
\end{table}

%and the accuracy of each method after one thousand evaluations for the CAR dataset.

%%%%%%%%%%%%%%%%%%%%%%%%%%%%%%%%%%%%%%%%%%
\section{Conclusion}
\label{section:conclusions}
%%%%%%%%%%%%%%%%%%%%%%%%%%%%%%%%%%%%%%%%%%
We presented the Ortho-MADS with the Nelder-Mead and the Variable Neighborhood Search (VNS), and the mesh size as a stopping criterion for tuning the hyperparameters $C$ and $\gamma$ of a SVM with Gaussian kernel. We have shown on benchmark datasets that the proposed approach outperforms widely used and state-of-the-art methods for model selection. The alternation between the search and pool stage provides a robust performance, and the use of two different search methods attenuate the randomness of the MADS and the starting point choice. Besides that, the proposed approach has convergence proof, and using the mesh size as stopping criterion gives to the user the possibility of setting a specific local minimum region instead of using the number of evaluations, time, or fixed-size grid as a stopping criterion. In the cases where a test set is available, the evolution of the mesh size during the BBO iterations can help identify under and over fitting behaviors.

The proposed approach gives the user the flexibility of choosing parameters to explore different strategies and situations. From our experiments, we recommend starting with the proposed default configuration, and from there the user can customize the Ortho-MADS parameters if necessary, which may improve the quality of the results. Therefore, we strongly recommend using the Ortho-MADS with NM and VNS search for tuning the hyperparameters of a  SVM with Gaussian kernel, which also provides many other functionalities. For future work, we expect to analyze the NOMAD with SVM variations, which includes incremental formulations and different kernels.

%% The Appendices part is started with the command \appendix;
%% appendix sections are then done as normal sections
%% \appendix

%% \section{}
%% \label{}
%\section*{References}

%%
%% Following citation commands can be used in the body text:
%% Usage of \cite is as follows:
%%   \cite{key}          ==>>  [#
%%   \cite[chap. 2]{key} ==>>  [#, chap. 2]
%%   \citet{key}         ==>>  Author [#]

%% References with bibTeX database:
\singlespacing 

%\bibliographystyle{model1-num-names}
%\bibliography{sample}

%% Authors are advised to submit their bibtex database files. They are
%% requested to list a bibtex style file in the manuscript if they do
%% not want to use model1-num-names.bst.

%% References without bibTeX database:

% \begin{thebibliography}{00}

%% \bibitem must have the following form:
%%   \bibitem{key}...
%%

% \bibitem{}

% \end{thebibliography}

\end{document}